%% file: main.tex
\documentclass[runningheads]{llncs}

\usepackage{eccv}

\usepackage{eccvabbrv}

\usepackage{graphicx}
\usepackage{booktabs}

\usepackage[accsupp]{axessibility}  

\usepackage{hyperref}

\usepackage{orcidlink}
\usepackage{algorithm} 
\usepackage{algorithmic}
\usepackage{multirow}
\newcommand{\yang}[1]{{\color{black}#1}}

\newcommand{\zhourixin}[1]{{\color{black}#1}}
\newcommand{\zhourixinTwo}[1]{{\color{black}#1}}

\definecolor{myblue}{RGB}{47,73,156} 
\definecolor{myorange}{RGB}{233,112,57} 
\definecolor{mygreen}{RGB}{2,125,63} 

\usepackage{fancyhdr} 
\fancypagestyle{firstpagefooter}{
	\fancyhf{} 
	\renewcommand{\headrulewidth}{0pt} 
	\renewcommand{\footrulewidth}{1pt} 
	\fancyfoot[L]{\tiny *Corresponding authors \\
 $\dagger$ Engineering Research Center of Knowledge-Driven Human-Machine Intelligence, MoE, China \\
 $\ddagger$ Key Laboratory of Ancient Chinese Script, Culture Relics and Artificial Intelligence, Jilin Province} 
}

\begin{document}

\title{PairingNet: A Learning-based Pair-searching and -matching Network for Image Fragments} 

\titlerunning{PairingNet}

\author{Rixin Zhou\inst{1}\orcidlink{0009-0005-2670-609X} \and
Ding Xia\inst{2}\orcidlink{0000-0002-4800-1112} \and
Yi Zhang\inst{3}\orcidlink{0000-0003-0294-8973} \and
Honglin Pang\inst{1}\orcidlink{0009-0005-4870-6605} \and
Xi Yang\inst{1,}\textsuperscript{*,$\dagger$, $\ddagger$}\orcidlink{0000-0001-5039-3680} \and
Chuntao Li\inst{4,}\textsuperscript{*, $\ddagger$}\orcidlink{0000-0001-9836-1493}}

\authorrunning{Zhou et al.}

\institute{School of Artificial Intelligence, Jilin University \and
The University of Tokyo \and
National University of Defense Technology, College of Electrionic Engineering\and 
School of Archaeology, Jilin University \\
\email{\{zhourx22\}@mails.jlu.edu.cn}
\email{\{lct33\}@jlu.edu.cn} \email{\{dingxia1995,panghlwork,earthyangxi\}@gmail.com}
\email{\{2356711993\}@qq.com}
}

\vspace{-0.6cm}
\maketitle

\pagestyle{headings}
\thispagestyle{firstpagefooter}


\begin{abstract}
   In this paper, we propose a learning-based image fragment pair-searching and -matching approach to solve the challenging restoration problem. Existing works use rule-based methods to match similar contour shapes or textures, which are always difficult to tune hyperparameters for extensive data and computationally time-consuming. Therefore, we propose a neural network that can effectively utilize neighbor textures with contour shape information to fundamentally improve performance. First, we employ a graph-based network to extract the local contour and texture features of fragments. Then, for the pair-searching task, we adopt a linear transformer-based module to integrate these local features and use contrastive loss to encode the global features of each fragment. For the pair-matching task, we design a weighted fusion module to dynamically fuse extracted local contour and texture features, and formulate a similarity matrix for each pair of fragments to calculate the matching score and infer the adjacent segment of contours. To faithfully evaluate our proposed network, we collect a real dataset and generate a simulated image fragment dataset through an algorithm we designed that tears complete images into irregular fragments. The experimental results show that our proposed network achieves excellent pair-searching accuracy, reduces matching errors, and significantly reduces computational time. Source codes and data are available at \href{https://github.com/zhourixin/PairingNet}{here}.
 
\end{abstract}

\begin{figure}[t]
    \centering
    \includegraphics[width=0.98\linewidth]{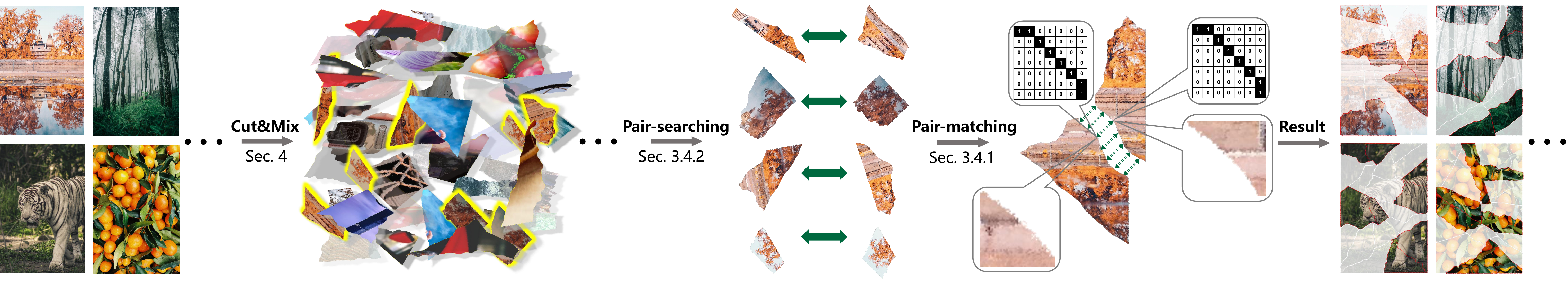}
    \captionof{figure}{Our pair-searching and -matching task. Given a large number of mixed image fragments, our proposed network can search and match pairs of fragments, which is important for restoration problems. We also designed an algorithm to generate a dataset of image fragments by tearing a set of complete images.}
  \label{fig1:examples}
\end{figure}

\input{sections/1_intro}
\input{sections/2_related}
\input{sections/3_method}
\input{sections/4_dataset}
\input{sections/5_exp}
\input{sections/6_conclusion}
\input{sections/7_supp}


%
%
\bibliographystyle{splncs04}
\bibliography{main}
\end{document}

%% file: sections/1_intro.tex
\vspace{-0.5cm}
\section{Introduction}
\label{sec:intro}
\vspace{-0.2cm}
Pair-searching and -matching of massive unorganized fragments is a critical problem in the research fields of computer vision and computer graphics, especially in the field of cultural relic restoration~\cite{da2002multiscale, huang2006reassembling, abitbol2021machine, funkhouser2011learning}. Traditional methods recover geometry transformation between fragments by calculating the similarity of contour segments based on geometric invariants such as curvature and manually designed descriptors~\cite{mikolajczyk2005performance, guo2016comprehensive}. It has a common core with the registration problem~\cite{brown1992survey}, which is to establish the correspondences of local features between related image pairs of the same scene. However, fragment assembly is more challenging because they do not have overlapping regions. Thus, existing rule-based methods are difficult to generalize to various complex situations. Furthermore, they are time-consuming since the computational complexity of finding matching pairs from a set of fragments is typically $O(N^2)$ in global search, and in practice, only a few pairs are hidden in a large number of mixed fragments.

Solving jigsaw puzzles~\cite{freeman1964apictorial, pomeranz2011fully, son2018solving} is also a similar task and learning-based approaches have been applied~\cite{paumard2018image, bridger2020solving}. However, two differences make these algorithms not generalizable. First, puzzle pieces usually have regular shapes, so they are best matched based on content information rather than contour features. Second, this problem usually studies matching all fragments into a complete block. Their input fragments do not come from mixed groups, and the matching results are dense and not lost. Therefore, they often use neighbor relationships to build graphs to constrain the search for the correct alignment of fragments. Once a fragment is lost, their closed-loop strategy fails.

In this paper, we propose PairingNet, an end-to-end deep learning network, for searching and matching pairs from a large number of image fragments. To address this problem, we first explore the use of Graph Convolutional Networks (GCNs) to extract more robust contour and texture features for our tasks. Second, we design a feature fusion way to learn local matching features by calculating the similarity matrix of each fragment pair. Third, we build a module based on the linear transformer to learn the global searching features for fragments using contrastive learning. Finally, we collect a real dataset and generate a image fragment dataset to provide data-driven and validate the effectiveness of our proposed network. The main contributions of our work are as follows:
\begin{itemize}
\setlength{\itemsep}{0pt}
\setlength{\parsep}{0pt}
\setlength{\parskip}{0pt}
    \vspace{-0.1cm}
    \item \yang{We propose a novel network to leverage the local contour and texture features for solving the pair-searching and -matching of image fragments. It provides an attempt to use advanced deep-learning techniques to solve this long-standing problem.}
    \item We provide a real dataset and a generated dataset with a carefully designed algorithm to simulate real-world fragments. The algorithm generates diverse image fragments to facilitate the application of subsequent learning-based methods.
    \item \yang{We conduct extensive experiments, ablation studies, and visual analysis to demonstrate the effectiveness of our proposed network.}

\end{itemize}

%% file: sections/2_related.tex
\vspace{-0.6cm}
\section{Related Work}
\vspace{-0.2cm}
\subsection{Fragments Assembly (Irregular Shape)}
Here, we focus on the pairwise matching algorithms of related representative works, although few of them also consider fragment relationships. Methods based on deep neural networks have been studied for object fragments; however, we could not find representative works for image fragments or even suitable datasets.
\vspace{-0.7cm}
\paragraph{Image fragments.}

Kong and Kimia~\cite{kong2001solving} solved fragment matching using geometric features (partial curve matching).
Leitao and Stolfi~\cite{da2002multiscale} described a specific multiscale matching method for the reassembly of 2D fragmented objects.
Tsamoura and Pitas~\cite{tsamoura2009automatic} presented a color-based method to automatically reassemble image fragments.
Huang et al.~\cite{huang2013mind} relied on salient curves detected inside the different image pieces to align gapped fragments.
Subsequently, both shape and appearance information along the boundaries were utilized and extracted for each piece for matching~\cite{liu2011automated, zhang2014graph}.
Furthermore, Derech et al.~\cite{derech2021solving} proposed to extrapolate each fragment to address fragment abrasion. 
For learning-based methods, traditional machine learning (SVM/Random Forest) was used as a classifier to process extracted local features~\cite{2013Learning, abitbol2021machine}. Then Le and Li~\cite{le2019jigsawnet} built a deep neural network to detect the compatibility of pairwise stitching.

\vspace{-0.2cm}
\paragraph{Object fragments.}

Huang et al.~\cite{huang2006reassembling} analyzed the geometry of the fracture surfaces by computing multi-scale surface characteristics.
Yang et al.~\cite{yang2017pairwise} matched stone tools based on contour points and their mean normals. Funkhouser et al.~\cite{funkhouser2011learning} learned a decision tree classifier to filter the candidates. Hong et al.~\cite{hong2021structure}  solved the problem of indistinguishable false matches by employing beam search to explore multiple registration possibilities. Ye et al.~\cite{ye2022puzzlefixer} presented an immersive system that can handle complex and ambiguous fragment shapes interactively with experts. 
In recent years, deep learning networks have been rapidly applied in 3D assembly.
Huang et al.~\cite{zhan2020generative} proposed an assembly-oriented dynamic graph learning framework to estimate the pose of 3D parts.
Chen et al.~\cite{chen2022neural} proposed Neural Shape Mating (NSM) to tackle pairwise 3D geometric shape mating.
Wu et al.~\cite{wu2023leveraging} leveraged SE(3) equivariance for such shape pose disentanglement. 

\vspace{-0.2cm}
\paragraph{Datasets.}

Existing image assembly works typically collect data in two ways: manual digitalization by photographing or scanning broken fragments, or using simple algorithms to tear complete images~\cite{2018Exploring, 2019A, 2013Learning}. The lack of a large-scale dataset prevents the emergence of deep learning networks for this task. However, multiple datasets have been introduced in 3D assembly problems~\cite{jones2021automate, sellan2022breaking, lamb2023fantastic}.

\vspace{-0.2cm}
\subsection{Jigsaw Puzzles (Regular Shape)}
\vspace{-0.1cm}
Solving jigsaw puzzles was first proposed in~\cite{freeman1964apictorial}, and then regular pieces became a key research object in computer vision. Many existing methods~\cite{cho2010probabilistic, pomeranz2011fully, gallagher2012jigsaw, son2014solving, son2018solving, gur2017square} focused on building relationships of pieces based on content information rather than contour-based pair-searching/-matching. And deep learning techniques~\cite{paumard2018image, bridger2020solving} was also used to solve this problem. 

%% file: sections/3_method.tex
\vspace{-0.4cm}
\section{Method}
\begin{figure*}[t]
\centering
    \includegraphics[width=1.0\textwidth]{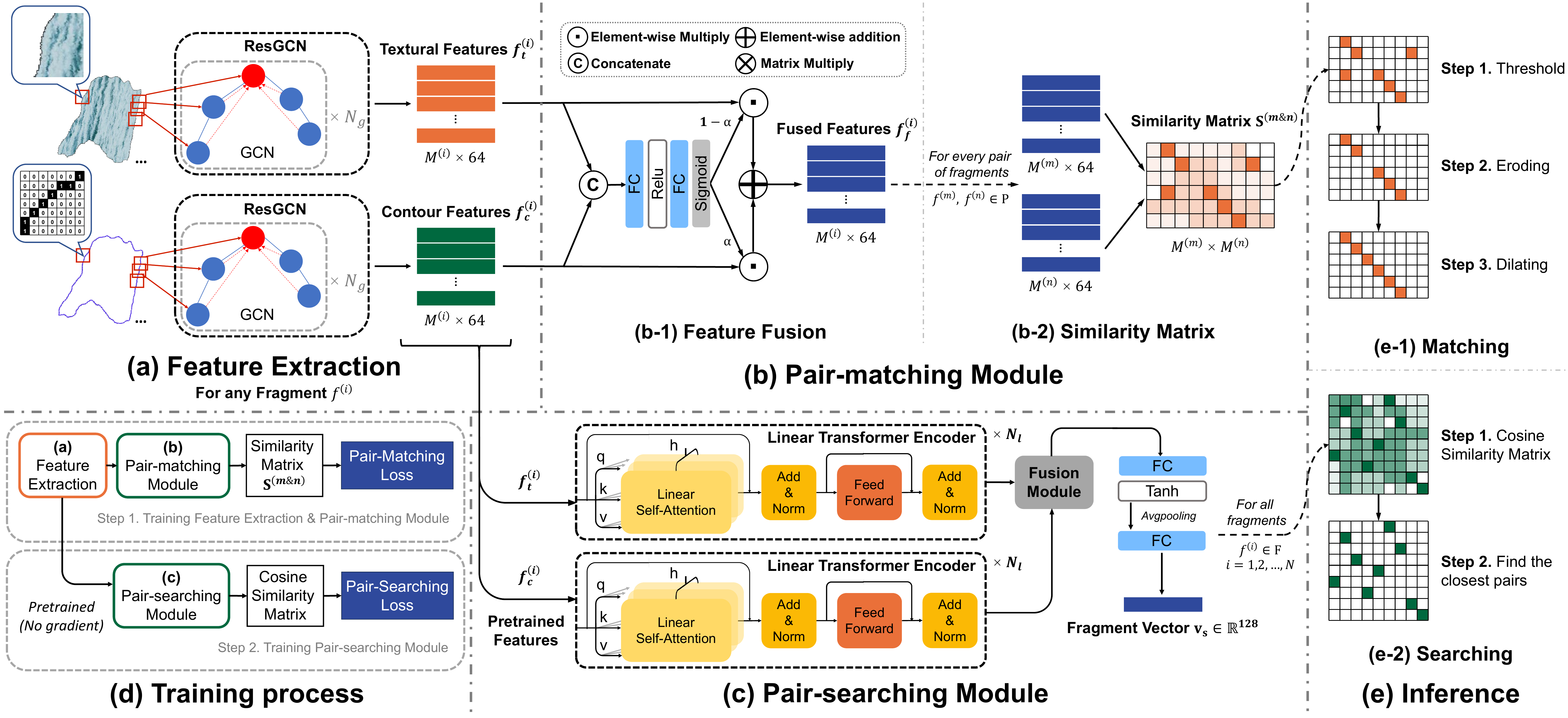}
    \caption{\yang{Pipeline of our proposed PairingNet. (a) Based on patches, we first employ a binary encoding to describe contours and use ResGCN as the backbone to extract the local contour and texture features of image fragments. (b) We design self-gated fusion to fuse the extracted features in a weighted manner to calculate the similarity matrix for each pair of fragments, and (c) a linear transformer-based encoder for learning the global features of each fragment. (d) We use a two-step strategy to train our proposed network. (e) During inference, we calculate cosine similarity to find the adjacent fragment pairs and further process the matching similarity matrix to establish better correspondences between a fragment pair.}}
    \label{fig:Stage 12 pipeline}
\end{figure*}

\vspace{-0.2cm}
\subsection{Problem Statement}
\vspace{-0.2cm}
Let us consider a fragment dataset $\mathbf{F}=\{f^{(i)}\}_{i=1}^{N}$, which consists of fragments $f$ torn from a certain image dataset. We define the contour points $c_j^{(i)} \in \mathbf{C}^{(i)}$ for each $f^{(i)}$, where $j = {1, 2, ..., M^{(i)}}$, $M^{(i)}$ is the contour length and $\mathbf{C}^{(i)}$ is an ordered set that contains all contour points of $f^{(i)}$. 


Regarding any two neighbor fragments sampled from a pairset $\mathbf{P}$, we denote the pair as $f^{(m)}$ and $f^{(n)}$. The \textit{pair-searching task} aims to retrieve the pairset $\mathbf{P}$ from a given fragment dataset $\mathbf{F}$. 
Meanwhile, for the contours of the paired fragments, $\mathbf{C}^{(m)}$ and $\mathbf{C}^{(n)}$, we can always find the overlapped part $\mathbf{C}^{(m\&n)}$. The \textit{pair-matching task} is designed to find the adjacent part from any given fragment pair.

\subsection{Network Architecture} 
\vspace{-0.1cm}
In the \textit{pair-matching task}, a model has the capability to discern differences between two fragments, regardless of whether these discrepancies are on a large or minimal scale. Simultaneously, the features extracted by the model can be used to quantify the likelihood of adjacency between any given two fragments, assisting in the identification of matched pairs. However, the process of comparing every two fragments is not only laborious, but also inefficient. Therefore, utilizing the rich features the \textit{matching model} extracts, we conceived a \textit{pair-searching} module. This module can adeptly convert any fragment into a unique feature vector, thereby significantly accelerating the pair-searching process.



\vspace{-0.1cm}
\subsubsection{Feature Extraction} 
\label{sec:Learning-based Matching Network}
\vspace{-0.3cm}
\paragraph{Contour encoding.}
\label{sec:Contour Encoding}
In this module, we aim to direct the model's attention to the unique features of contours. We design distinct contour encodings that also may be applied in other sketch-based fields. We tested various patch sizes and all these distinct encodings, finalizing a binary encoding and a patch size of 7x7 for our network. Subsequently, these contour patches are processed through a convolutional layer, followed by an average pooling layer and a fully connected network. We then adopt ResGCN~\cite{li2019deepgcns} to amalgamate adjacent features, further extracting comprehensive global features. As such, for any provided fragment input denoted as $f^{(i)}$ with its corresponding contour $\mathbf{C}^{(i)}$, this module yields a return of $f_c^{(i)} \in \mathbb{R}^{M^{(i)} \times 64}$.


\vspace{-0.3cm}
\paragraph{Textural encoding.}
\label{sec:textual_feature}
Our objective is to accentuate the distinguishing characteristics of textures in this module. We begin by cropping the image using a patch centered on each contour point. This is followed by filtering these patches through two consecutive convolutional layers, coupled with a single average pooling operation. Then, a ResGCN~\cite{li2019deepgcns} is integrated to compile adjacent features, thereby enhancing the receptive domain of local textural characteristics. For any given fragment input represented as $f^{(i)}$ along with its associated contour $\mathbf{C}^{(i)}$, this module produces an output of $f_t^{(i)} \in \mathbb{R}^{M^{(i)} \times 64}$.

\vspace{-0.1cm}
\subsubsection{Pair-matching Module}
\vspace{-0.4cm}
\paragraph{Feature Fusion.}
\label{sec:fusion}
Regarding the decision-making process for pair-matching and -searching tasks, the dependence on contour and textural features varies. For instance, in scenarios where both fragments are solid colors, texture serves as the primary identifying factor suggesting adjacency. However, texture alone cannot identify the matched contours. In contrast, simple contours, like straight lines, require texture features for accurate matching position determination, as contour information alone falls short.

Given this, we propose the incorporation of a \textit{Feature Fusion} module~\cite{dai2021attentional} which would be capable of adaptively fine-tuning the weights assigned to these two features. The specific structure of this integral component is depicted in Figure~\ref{fig:Stage 12 pipeline} (b).
First, we concatenate $f_t^{(i)}$ and $f_c^{(i)}$, then the weight $\mathbf{w}^{(i)}$ for each contour point $c_j^{(i)} \in \mathbf{C}^{(i)}$ is computed by:
\begin{equation}
    \small
    \mathbf{w}^{(i)} = \sigma(\mathbf{W}_g(f_t^{(i)} \textcircled{c} f_c^{(i)}) + \mathbf{b}_g)^{(i)}
\end{equation}
where $\sigma$ is $sigmoid$ the activation function, $\textcircled{c}$ means channel-wise concatenation, and $\mathbf{w} \in \mathbb{R}^{M^{(i)} \times 64}$.
Second, we fuse features of two modalities:
\begin{equation}
    \small
    \mathbf{F}_f^{(i)} = \mathbf{w}^{(i)} \odot \mathbf{F}_t^{(i)} + (1-\mathbf{w}^{(i)}) \odot \mathbf{F}_c^{(i)}
\end{equation}
where $\mathbf{F}_f^{(i)} \in \mathbb{R}^{M^{(i)} \times 64}$, and $\odot$ is element-wise multiplication.
With this module, we let the model adaptively decide the weights between textual and contour features, which is proven effective in later experiments.

\vspace{-0.2cm}
\paragraph{Similarity Matrix.}
\label{sec:sim_matrix}
\zhourixin{As shown in Figure~\ref{fig:Stage 12 pipeline} (b-2)}, for every pair of fragments $f^{(m)}$ and $f^{(n)}$ from the pairset $\mathbf{P}$ with contour points $\mathbf{C}^{(m)}$ and $\mathbf{C}^{(n)}$, we are able to formulate the similarity matrix $\mathbf{S}^{(m\&n)} \in \mathbb{R}^{M^{(m)} \times M^{(n)}}$ ~\cite{dual-softmax, rocco2018neighbourhood}. 

We first calculate the similarity matrix $\tilde{\mathbf{S}}^{(m\&n)}$ by matrix multiplication:
\begin{equation}
    \small
    \tilde{\mathbf{S}}^{(m\&n)} = \frac{1}{\sqrt{D}} \mathbf{F}_f^{(m)} (\mathbf{F}_f^{(n)})^T
\end{equation}
where $D$ is the dimension of the feature, in our model, $D$ is 64.
Then, each entry in the matrix $\mathbf{S}^{(m\&n)}$ is normalized:
\begin{equation}
    \small
    \mathbf{S}^{(m\&n)}(i, j) =  \frac{exp(\tilde{\mathbf{S}}^{(m\&n)}_{i, j})}{\sum_i{exp(\tilde{\mathbf{S}}^{(m\&n)}_{i, j})}} \frac{exp(\tilde{\mathbf{S}}^{(m\&n)}_{i, j})}{\sum_j{exp(\tilde{\mathbf{S}}^{(m\&n)}_{i, j})}}
\end{equation}

\vspace{-0.2cm}
\subsubsection{Pair-searching Module}

We introduce a modified feature fusion module to process features extracted by the backbone network and encode any fragment $f^{(i)}$ into a feature vector $v_s^{(i)} \in \mathbb{R}^{128}$. It is based on the discovery that using fused features in the pair-matching task will lower the performance. 

To be specific, \zhourixin{as shown in Figure~\ref{fig:Stage 12 pipeline} (c)}, every feature branch utilizes a Linear Transformer Encoder~\cite{wang2020linformer} to enhance the encoding of both contour and textural features. Subsequent to the modified fusion module, we add a fully-connected layer to encode the fused feature regarding per contour point. Then an average pooling layer is applied to map the feature matrix to a singular vector. Finally, another fully-connected layer is used to encode the feature.

For every pair of fragment $f^{(i)}$ and $f^{(j)}$ from the fragment set $\mathbf{F}$, this module allows them to be encoded as $v_s^{(i)}$ and $v_s^{(j)}$. By calculating their inner product, we can efficiently quantify and contrast their similarity, thereby expediently identifying neighbor pairs.


\vspace{-0.2cm}
\subsection{Loss Function} 
\vspace{-0.2cm}
\zhourixin{We use two loss functions to train our network, and the training process is shown in Figure~\ref{fig:Stage 12 pipeline} (d).}
\subsubsection{Pair-matching Loss}
Given the extreme imbalance in the ratio of positive to negative samples in the pair-matching task, we adopt the concept of focal loss~\cite{li2022lepard} to concentrate on learning challenging correspondences.

For any paired fragments $f^{(m)}$ and $f^{(n)}$, entries in the ground truth similarity matrix $\mathbf{S}_{gt}^{(m\&n)}$ are 1 for matched contour points and 0 for unmatched ones. We calculate the matching loss $\mathcal{L}_{match}^{(m\&n)}$ as follows:

\vspace{-0.2cm}
\begin{equation}
\small
\begin{aligned}
    \mathcal{L}_{match}^{(m\&n)} & = -\sum(\beta_1 (\textbf{1}-\mathbf{S}^{(m\&n)})^\gamma \log {\mathbf{S}^{(m\&n)}} \mathbf{S}^{(m\&n)}_{gt} \\ 
    & + \beta_2 (\mathbf{S}^{(m\&n)})^\gamma \log{(\textbf{1}-\mathbf{S}^{(m\&n)})} (\textbf{1}-\mathbf{S}^{(m\&n)}_{gt})) \\
\end{aligned}
\end{equation}
\vspace{0.03cm}

\noindent where $\beta_1$ and $\beta_2$ are the balancing weights and $\beta_1 + \beta_2 = 1$, $\gamma$ is the decay factor. $\sum$ adds all the entries in the input matrix and all operations except $\sum$ are element-wise. By calculating and adding the pair-matching loss for all paired fragments $f^{(m)}, f^{(n)}$ from the pairset $\mathbf{P}$, we get $\mathcal{L}_{match}$.


\vspace{-0.3cm}
\subsubsection{Pair-searching Loss}
In the pair-searching task, we employ InfoNCE loss~\cite{oord2018representation} as the contrastive loss function $\mathcal{L}_{search}$. We utilize paired samples for constructing positive sample pairs, and unpaired samples for negative sample pairs.


\vspace{-0.2cm}
\subsection{Inference}
\label{sec:inference}
\vspace{-0.2cm}
\subsubsection{Pair-matching}\label{sec:contour_inf}
In this section, we target to retrieve matched contour from the similarity matrix $\mathbf{S}^{(m\&n)}$.

We first apply a threshold $\epsilon = 0.006$ to filter out noisy entries. Entries with value lower than the threshold is set to 0.
Because contour pointset $\mathbf{C}^{(m)}$ and $\mathbf{C}^{(n)}$ are ordered sets, for any ground truth similarity matrix $\mathbf{S}^{(m\&n)}_{gt}$, if the start for contour points is the left-top corner,
we can observe that the part with a value of 1 in the matrix must be arranged in the direction from upper left to lower right and connected into a line segment. Therefore, this inspires us to enhance the filtered similarity matrix. We design an eroding kernel $\mathbf{K}_e$ to filter out not dominant segments, which is adopted once. Then a dilating kernel $\mathbf{K}_d$ is applied once to connect close but split segments. The definitions of two kernels are as follows:




\begin{equation}
    \small
\mathbf{K}_{e} = \begin{bmatrix}
0 & 0 & 1 \\
0 & 0 & 0 \\
1 & 0 & 0
\end{bmatrix}, \quad
\mathbf{K}_{d} = \begin{bmatrix}
0 & 0 & 1 \\
0 & 1 & 0 \\
1 & 0 & 0
\end{bmatrix}.
\end{equation}




Finally, we use RANSAC~\cite{fischler1981random} to estimate the correspondences of two fragments and obtain the transformation matrix.


\vspace{-0.3cm}
\subsubsection{Pair-searching} \label{sec:sim_score_cal}

For N fragments $f^{(i)} \in \mathbf{F}, i=1,2,...,N$, using the searching module, we can get a feature matrix $\mathbf{V}_s=[v_1,v_2,\ldots,v_N]^T \in \mathbb{R}^{N \times 128}$.
Subsequently, we compute the cosine similarity between all global features to form the cosine similarity matrix. This allows us to obtain the similarity score for any pair of fragments.

%% file: sections/4_dataset.tex
\vspace{-0.4cm}
\section{Dataset}
\vspace{-0.4cm}
\begin{figure}[t]
    \centering
    \includegraphics[width=0.98\linewidth]{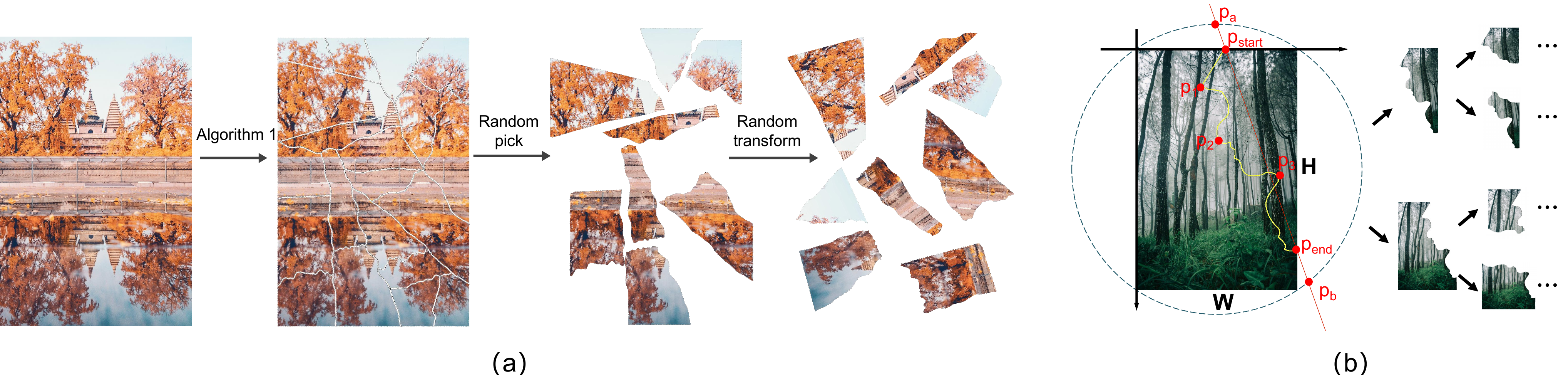}
    \caption{Dataset creation. (a) The process of generating our dataset. (b) Illustration of Algirithm~\ref{alg:data generate} for tearing a complete image.}
    \label{fig:cut result and line}
\end{figure}

\begin{table}[t]
\centering
\caption{Dataset division. We divide the entire test set into three difficulty levels according to the contour overlapping proportion of fragment pairs.}
\label{Table of data count}
\resizebox{0.6\linewidth}{!}{ 
\begin{tabular}{c|c|c|c|cccc}
\hline
\multirow{2}{*}{} & \multirow{2}{*}{Total} & \multirow{2}{*}{Train} & \multirow{2}{*}{Validation} & \multicolumn{4}{c}{Test}                                                                   \\ \cline{5-8} 
                  &                        &                        &                             & \multicolumn{1}{c|}{Full} & \multicolumn{1}{c}{High} & \multicolumn{1}{c}{Medium} & Low  \\ \hline\hline
Fragments         & 8196                   & 4098                   & 819                         & \multicolumn{1}{c|}{3279} & \multicolumn{1}{c}{1071} & \multicolumn{1}{c}{1671}   & 1090 \\ 
Pairs             & 14951                      & 3654                   & 137                         & \multicolumn{1}{c|}{2370} & \multicolumn{1}{c}{663}  & \multicolumn{1}{c}{1103}   & 604  \\ 
\hline
\end{tabular}
}
\end{table}

\begin{algorithm}[t]
    \scriptsize
	\renewcommand{\algorithmicrequire}{\textbf{Input:}}
	\renewcommand{\algorithmicensure}{\textbf{Output:}}
	\begin{algorithmic}[1]
        \REQUIRE {List of fragments $L_{frag}=\{\}$, a complete image $I$};
        \\
        Max iterations $t_{max}=40$,
        min arc length ratio $\tau=0.9$, max number of cutting points $N_{max}=3$, min distance from cutting point to edge of fragment $D_{min}=100$, 
        the number of Fourier orthogonal basis $n=20$, scaling ratios $s_1=0.25$, $s_2=0.0067$, $s_3=1.5$, $s_4=0.3$, 
        probability of curve type $\rho=0.5$, min fragment size $h_{min}=150$, $w_{min}=150$.
 
        \ENSURE  {Collection of fragments $L_{frag}$}

        \STATE put $I$ in $L_{frag}$ 
        \WHILE {$t < t_{max}$}
            \STATE $f\gets$randomly\_pick$(L_{frag})$, $R\gets$bounding\_rectangle$(f)$
            \STATE $W,H\gets$width\_and\_height$(R)$, $C\gets$circumcircle$(R)$
            \STATE $l_{dia}\gets$diameter$(C)$
            \STATE
            \STATE // generate the cutting points
            \REPEAT
                \STATE $p_a$, $p_b$ $\gets$ randomly\_pick$(C)$
                \STATE $l_{arc}\gets$smaller\_arc\_length$(p_a, p_b)$
            \UNTIL {$l_{arc} > \tau \cdot half\_perimeter(C)$}
            \STATE $p_{start}$, $p_{end}\gets$ intersection$(f, p_a, p_b)$
            
            \STATE $m\gets$randomly\_pick$\{0, \ldots, N_{max}\}$
            \REPEAT
                \STATE [$p_1$, $\ldots$, $p_m$] $\gets$ random\_select\_inside\_points$(f, m)$
                \STATE $d\gets$min\_distance\_to\_edge($f, p_1$, $\ldots$, $p_m$)
            \UNTIL {$d > D_{min}$}
            \STATE put $p_{start}$, $p_{end}$ in [$p_1$, $\ldots$, $p_m$]
            
            \STATE
            \STATE // generate the cutting segments
            \FOR{each $p_i \in $ [$p_{start}$, $p_1$, $\ldots$, $p_m$, $p_{end}$]}
                \STATE irregular line $l_{irr}=\{\}$, phase $\phi \sim \mathcal{U}(-\pi,\pi)$, \\
                amplitude $A\sim\mathcal{N}(s_1H, s_2H)$, period $T \sim \mathcal{N}(s_3W, s_4W)$
                \FOR{each $x$-coordinate between $(p_i, p_{i+1})$}
                    \STATE $y \gets \sum_{i=0}^n \frac{A}{1+i}sin(\frac{2\pi i}{T} x+\phi)+\frac{H}{2}$
                    \STATE put $\{x, y\}$ in $l_{irr}$
                \ENDFOR
            \STATE $l_{str} \gets$ straight\_line$(p_i, p_{i+1})$
            \STATE $l_{cut} \gets$ select\_and\_connect$(l_{irr}, l_{str}, \rho, l_{cut})$
            \ENDFOR
            
            \STATE $f_1$, $f_2 \gets$ cut\_fragment$(f, l_{cut})$
            
            \IF {size$(f_1)>h_{min}, w_{min}$ and size$(f_2)>h_{min}, w_{min}$}
                \STATE put $f_1$, $f_2$ in $L_{frag}$ 
            \ENDIF
         \ENDWHILE

		
	\end{algorithmic}  
\captionsetup{font=scriptsize}
\caption{Generate fragments of an image}
\label{alg:data generate}
\end{algorithm}
\vspace{-0.2cm}
\paragraph{Generated Dataset.}
\vspace{-0.2cm}
We design Algorithm~\ref{alg:data generate} to tear a series of complete images and create a new image fragment dataset with data close to the real world. The process is also shown in Figure~\ref{fig:cut result and line}. 
To control the shape and area of the fragments, we first find the circumscribed circle of a fragment and take two points at an appropriate distance on the circle to determine the starting and ending points of the segmentation. Then, to simulate a more realistic fragment contour, we randomly divide the cutting line into several segments, each segment randomly choosing a straight line or an irregular curve. The irregular curve is synthesized by a series of Fourier orthogonal bases to make the shape of the fragments more diverse and similar to the edge of a real torn paper. We set a minimum threshold for the fragments, and the fragment is re-segmented if a very small fragment appears.


\begin{figure}[t]
\centering
    \includegraphics[width=1\linewidth]{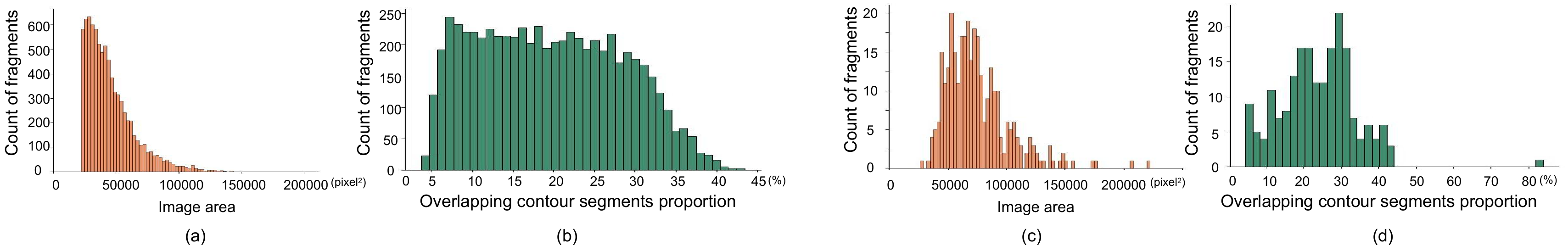}
    \caption{Dataset statistics. (a) Image area distribution of the generated fragments. (b) Proportional distribution of overlapping contour segments of the generated fragments. \zhourixin{(c) Image area distribution of the scanned real fragments. (b) Proportional distribution of overlapping contour segments of the scanned real fragments.}}
    \label{fig:data distribution}
\end{figure}

To create our dataset, we collect $390$ images of ten themes from Pexels~\cite{pexel} and empirically set the generation parameters shown in Algorithm~\ref{alg:data generate} to obtain fragments. We calculate the distributions of our generated data, as shown in Figure~\ref{fig:data distribution}. The distribution of fragment areas exhibits a long-tail pattern, which is consistent with the distribution of fragment areas observed when tearing large amounts of paper in real-life scenarios. The contour overlapping proportion of fragment pairs is closer to a uniform distribution over most ranges, indicating that our algorithm generates various types of fragment pairs. Then, we divide the dataset into training set, validation set, and test set in 5:1:4. We further divide the full test set into three difficulty levels based on the overlapping proportion of contour segments between fragment pairs. \yang{Low difficulty means that the pair of fragments has larger overlapping contour segments, and vice versa.} The details are shown in Table~\ref{Table of data count}. 

\paragraph{Real Dataset.}
\zhourixin{We also collect a series of real image fragments to construct a real dataset. We printed 34 complete images (not included in our generated dataset), manually cut them into fragments, and then scanned them. Then, we apply some post-processing steps to the scanned image fragments to obtain the contour of the fragments and the annotation of the matching points. More details about real dataset can be found in the supplementary materials.}

%% file: sections/5_exp.tex
\vspace{-0.2cm}
\section{Experiments}
\vspace{-0.2cm}
\subsection{Comparison Methods}
\vspace{-0.1cm}

Finding comparison methods is not easy. Because rule-based methods have been around for quite some time, lack source code, and tend to exhibit lower performance compared to advanced deep learning methods. However, most existing deep learning methods require extensive modifications if applied to our task. Therefore, we employ Jigsawnet~\cite{le2019jigsawnet} (without loop consistency) and a rule-based method~\cite{zhang2014graph} (Step 1) as our comparison methods.

\subsection{Implementation Details}

\yang{Our PairingNet was implemented by PyTorch, and conducted on a server with 4 RTX A40 GPUs and Intel$\circledR$ Xeon$\circledR$ Gold 5220 CPUs (72 cores).} 
We used the Adam optimizer with an initial learning rate of $0.001$, which was adjusted using a cosine annealing strategy. 
For training pair-matching module, the batch size was set to $20$, and for the pair-searching module, the batch size was set to $175$. 

\zhourixin{For comparison methods, given that the performance bottleneck of them resides in the CPU, utilizing GPUs for computational acceleration is deemed inappropriate. Thus we use two more powerful CPUs computing platforms to run the comparison method.} We used the official sourcecode (implemented by TensorFlow) of Jigsawnet and conducted experiments on a server with 4 Quadro RTX 6000 GPUs and Intel$\circledR$ Xeon$\circledR$ Gold 5220 CPUs (144 cores). We reproduced the rule-based method by Python and conducted experiments on a node of a Supercomputing platform with Intel$\circledR$ Xeon$\circledR$ Gold 6258R CPUs (128 cores). 

For evaluation metrics, we employ Recall@k and NDCG@k~\cite{busa2012apple} for the pair-searching task, and Registration recall (RR), Hausdorff distance (HD), Radians Error (RE) and Normalized Translation Error (NTE) for the pair-matching task. \yang{We normalize Translation Error (TE) by the sum of the areas of fragment pairs to obtain the NTE, considering the large difference in area between fragments.}




\subsection{Results}

\begin{table*}[htbp]
\centering
\footnotesize
\caption{Comparison Results of various metrics. The best results in our full dataset are highlighted in bold font, and the best results for each difficulty are marked in different colors.}
\label{Table of comparable method}
\resizebox{1\textwidth}{!}{ 
\begin{tabular}{c|c|c|c|c|c|c|c|c|c|c|c}
\hline
Method                       & Difficulty & Recall@5                              & Recall@10                             & Recall@20                             & NDCG@5                                & NDCG@10                               & NDCG@20                               & RR $\uparrow$            & HD $\downarrow$           & RE $\downarrow$          & NTE $\downarrow$              \\ \hline\hline
                             & High       & 0.261                                 & 0.354                                 & 0.443                                 & 0.185                                 & 0.218                                 & 0.241                                 & 0.107                                 & 212.003                                & 1.289                                 & 37.100$\times 10^{-4}$                                 \\
                             & Medium     & 0.329                                 & 0.427                                 & 0.529                                 & 0.242                                 & 0.275                                 & 0.302                                 & 0.153                                 & 205.463                                & 1.249                                 & 33.742$\times 10^{-4}$                                 \\
                             & Low        & 0.358                                 & 0.464                                 & 0.555                                 & 0.261                                 & 0.296                                 & 0.32                                  & 0.188                                 & 210.274                                & 1.231                                 & 32.692$\times 10^{-4}$                                 \\ \cline{2-12} 
\multirow{-4}{*}{Rule-based~\cite{zhang2014graph}} & All        & 0.317                                 & 0.416                                 & 0.511                                 & 0.248                                 & 0.284                                 & 0.311                                 & 0.149                                 & 208.519                                & 1.256                                 & 34.414$\times 10^{-4}$                                 \\ \hline\hline
                             & High       & {\textcolor{myorange}{ \textbf{0.305}}} & 0.394                                 & 0.478                                 & {\textcolor{myorange}{ \textbf{0.231}}} & {\textcolor{myorange}{ \textbf{0.261}}} & {\textcolor{myorange}{ \textbf{0.284}}} & 0.117                                 & 226.026                                & 1.306                                 & 37.011$\times 10^{-4}$                                 \\
                             & Medium     & 0.383                                 & 0.478                                 & 0.565                                 & 0.309                                 & 0.34                                  & 0.364                                 & 0.213                                 & 206.040                                & 1.155                                 & 31.467$\times 10^{-4}$                                 \\
                             & Low        & 0.459                                 & 0.548                                 & 0.631                                 & 0.368                                 & 0.397                                 & 0.419                                 & 0.241                                 & 202.942                                & 1.186                                 & 30.500$\times 10^{-4}$                                 \\ \cline{2-12} 
\multirow{-4}{*}{Jigsawnet~\cite{le2019jigsawnet}}  & All        & 0.38                                  & 0.472                                 & 0.557                                 & 0.326                                 & 0.359                                 & 0.383                                 & 0.194                                 & 210.842                                & 1.205                                 & 32.772$\times 10^{-4}$                                 \\ \hline\hline
                             & High       & 0.276                                 & {\textcolor{myorange}{ \textbf{0.413}}} & {\textcolor{myorange}{ \textbf{0.535}}} & 0.189                                 & 0.234                                 & 0.268                                 & {\textcolor{myorange}{ \textbf{0.606}}} & {\textcolor{myorange}{ \textbf{92.229}}} & {\textcolor{myorange}{ \textbf{0.807}}} & {\textcolor{myorange}{ \textbf{18.335$\times 10^{-4}$}}} \\
                             & Medium     & {\textcolor{mygreen}{\textbf{0.501}} } & {\textcolor{mygreen}{\textbf{0.619}} } & {\textcolor{mygreen}{ \textbf{0.732}}} & {\textcolor{mygreen}{ \textbf{0.382}}} & {\textcolor{mygreen}{ \textbf{0.423}}} & {\textcolor{mygreen}{\textbf{0.454}} } & {\textcolor{mygreen}{\textbf{0.879}} } & {\textcolor{mygreen}{\textbf{35.617}} } & {\textcolor{mygreen}{\textbf{0.236}} } & {\textcolor{mygreen}{\textbf{5.685$\times 10^{-4}$}} }  \\
                             & Low        & {\textcolor{myblue}{ \textbf{0.717}}} & {\textcolor{myblue}{ \textbf{0.805}}} & {\textcolor{myblue}{ \textbf{0.879}}} & {\textcolor{myblue}{ \textbf{0.581}}} & {\textcolor{myblue}{ \textbf{0.61}}}  & {\textcolor{myblue}{ \textbf{0.629}}} & {\textcolor{myblue}{ \textbf{0.961}}} & {\textcolor{myblue}{ \textbf{12.336}}} & {\textcolor{myblue}{ \textbf{0.09}}}  & {\textcolor{myblue}{ \textbf{1.936$\times 10^{-4}$}}}  \\ \cline{2-12} 
\multirow{-4}{*}{Ours}       & All        & \textbf{0.493}                        & \textbf{0.608}                        & \textbf{0.714}                        & \textbf{0.417}                        & \textbf{0.458}                        & \textbf{0.487}                        & \textbf{0.835}                        & \textbf{43.116}                        & \textbf{0.352}                        & \textbf{7.735$\times 10^{-4}$}                         \\ \hline
\end{tabular}

}
\end{table*}

\begin{figure}[t]
\centering
    \includegraphics[width=0.98\linewidth]{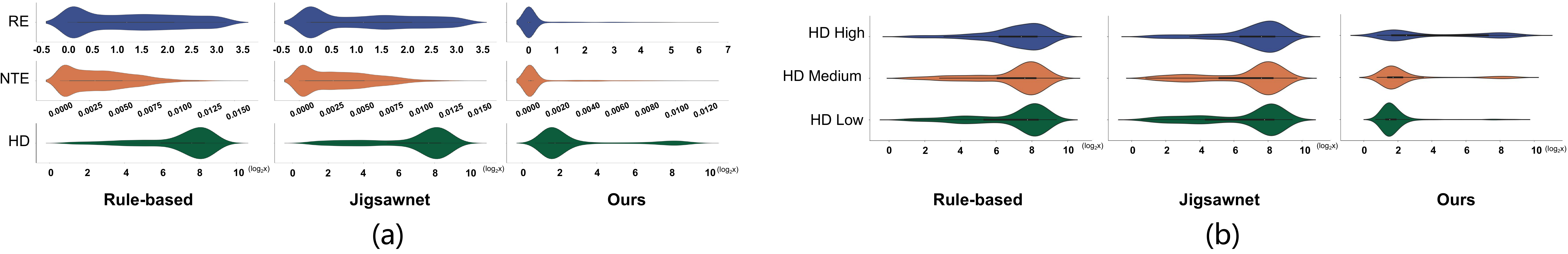}
    \caption{Violin plots of the matching results of our method and the comparison method: (a) RE, NTE, and HD on the test set. (b) HD on the test sets of different difficulties.}
    \label{fig:HD and RE}
\end{figure}


\begin{figure}[t]
\centering
    \includegraphics[width=0.98\linewidth]{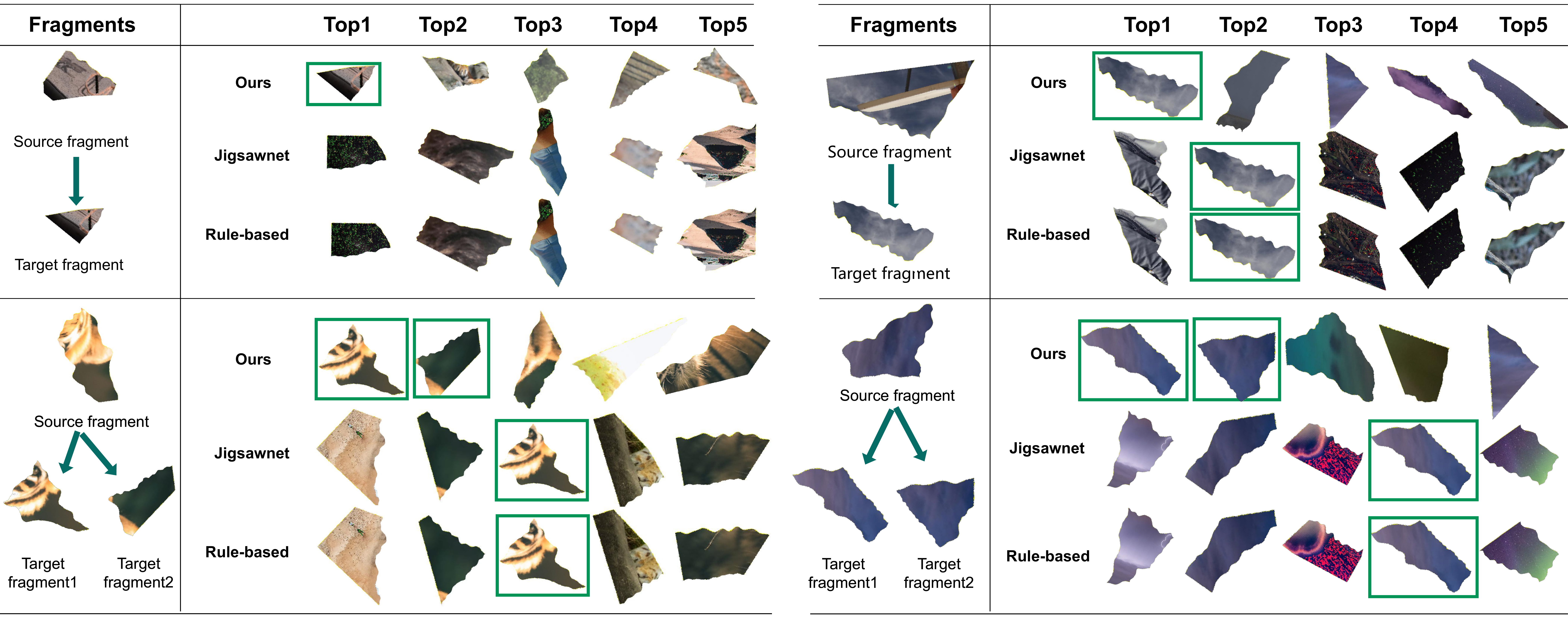}
    \caption{Examples of pair-searching results. Our network is able to accurately identify the corresponding target fragment (highlighted with a green box).}
    \label{fig:searching comparision}
\end{figure}

\begin{figure}[t]
\centering
    \includegraphics[width=1\linewidth]{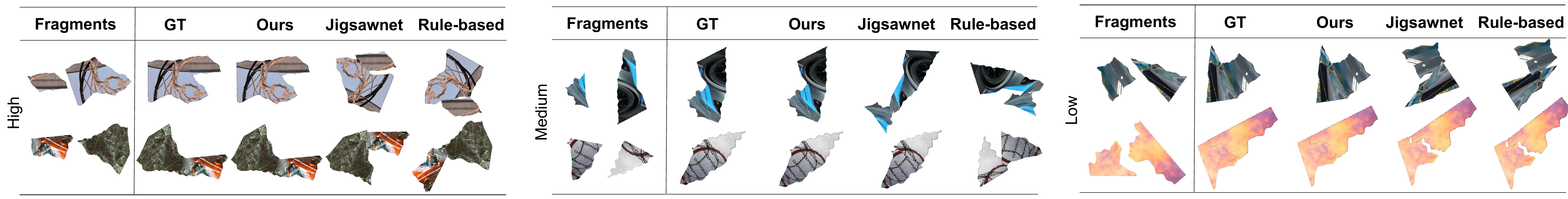}
    \caption{\yang{Examples of pair-matching results for three difficulty levels. Our proposed network achieves satisfactory matching results. However, for low-difficulty cases, Jigsawnet and the Rule-based method find approximate matching positions, while the matchings are not accurate enough. For medium- and high-difficulty cases, both comparison methods are difficult to obtain correct matching results.}}
    \label{fig:comparision}
\end{figure}

\begin{table}[t]
\centering
\caption{\zhourixin{Comparison Results of various metrics on our real dataset. The best results are highlighted in bold font.}}
\label{Table of real dataset comparison}
\resizebox{0.90\linewidth}{!}{ 
\begin{tabular}{c|c|c|c|c|c|c|c|c|c|c}
\hline
Method     & Recall@5       & Recall@10      & Recall@20      & NDCG@5         & NDCG@10        & NDCG@20        & RR $\uparrow$ & HD $\downarrow$ & RE $\downarrow$ & NTE $\downarrow$ \\ \hline\hline
Rule-based~\cite{zhang2014graph} & 0.138          & 0.247          & 0.366          & 0.079          & 0.114          & 0.144          & 0.128                      & 7.698                        & 1.614                        & 30.746$\times 10^{-4}$                   \\ \hline
Jigsawnet~\cite{le2019jigsawnet}  & 0.237          & 0.326          & 0.425          & 0.185          & 0.213          & 0.238          & 0.153                      & 7.570                        & 1.545                        & 29.061$\times 10^{-4}$                   \\ \hline
Ours       & \textbf{0.415} & \textbf{0.564} & \textbf{0.678} & \textbf{0.321} & \textbf{0.369} & \textbf{0.399} & \textbf{0.752}             & \textbf{4.182}               & \textbf{0.480}               & \textbf{9.075$\times 10^{-4}$}           \\ \hline
\end{tabular}

}
\end{table}

\begin{figure}[t]
\centering
    \includegraphics[width=0.98\linewidth]{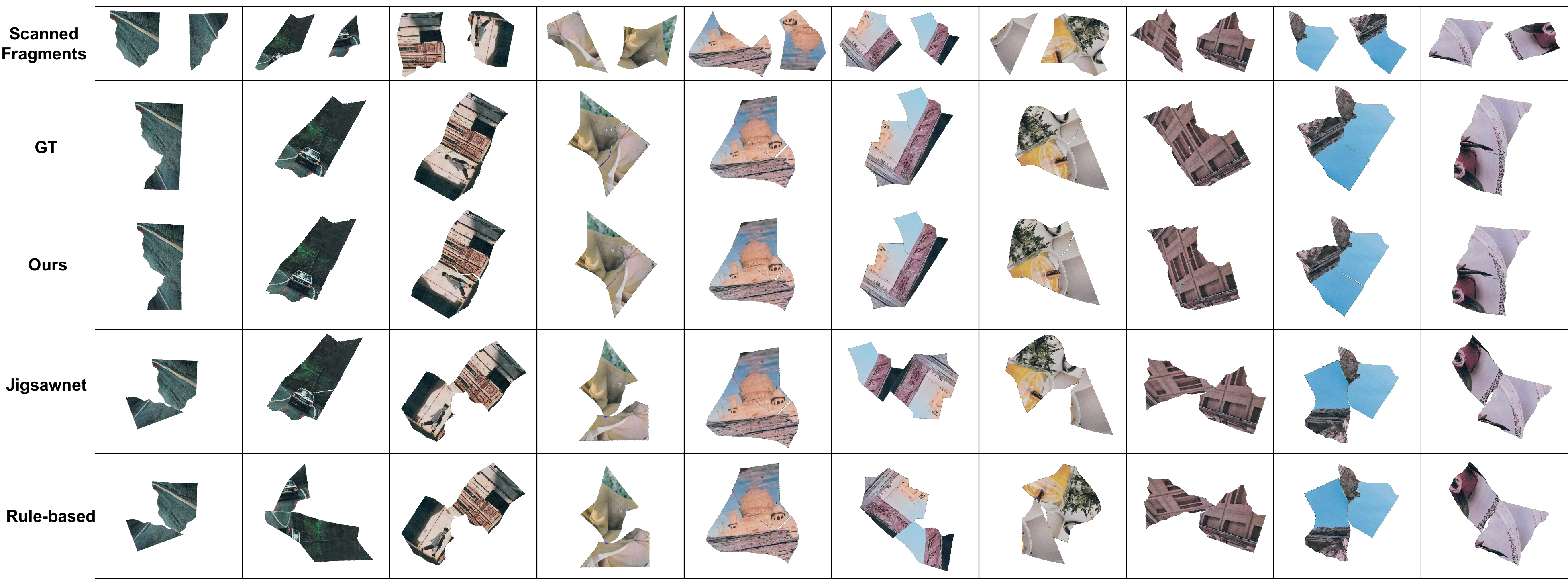}
    \caption{\zhourixin{Examples of pair-matching results of PairingNet on real dataset. Without retraining, our method can also effectively match the scanned real fragments.}}    
    \label{fig:real matching}
\end{figure}

\begin{figure}[t]
\centering
    \includegraphics[width=0.53\linewidth]{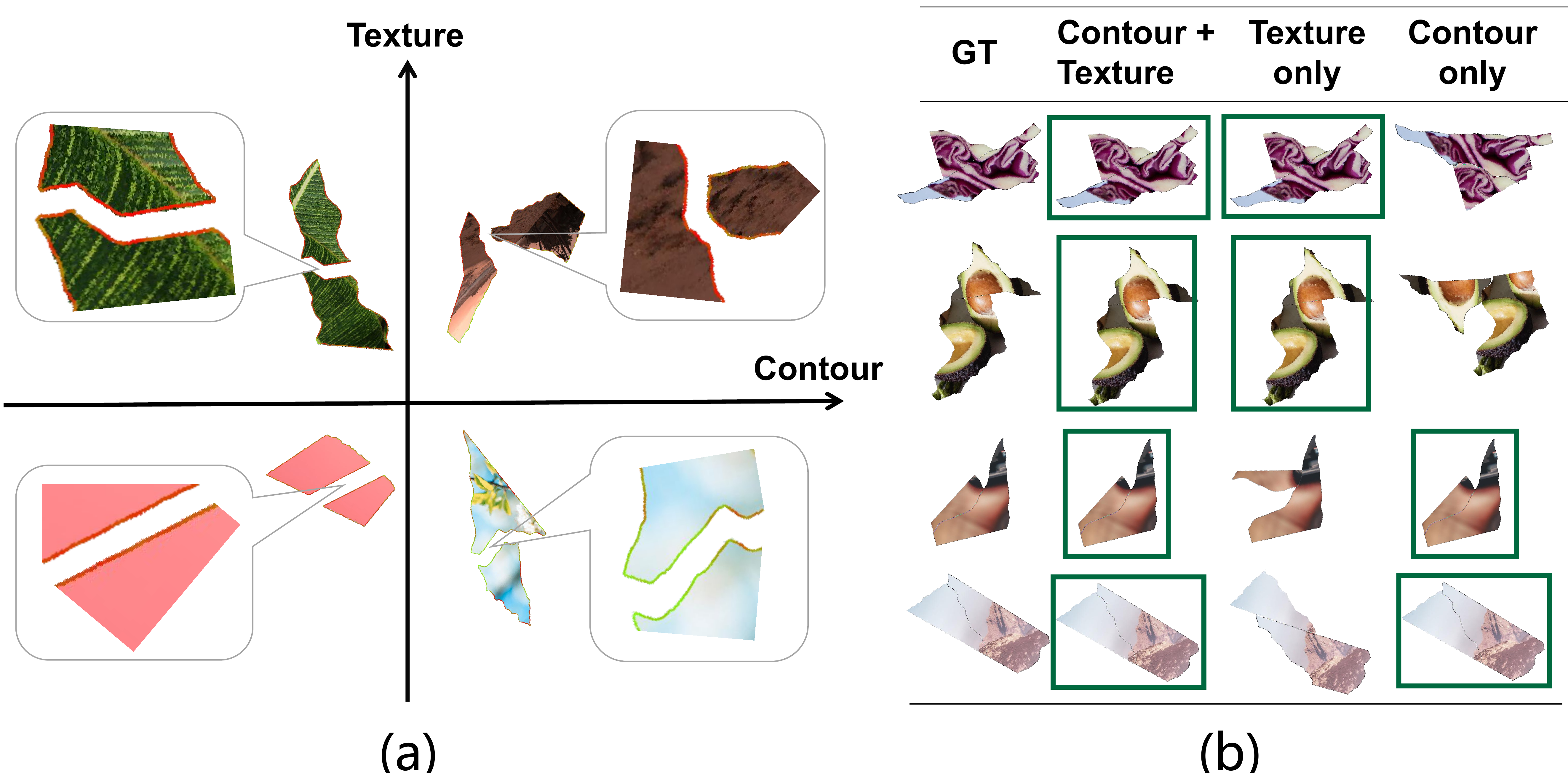}
    \caption{\yang{Visual analysis of Texture/Contour. (a) We visualize the weights of texture and contour features in our network, with colors closer to red representing more use of texture features and closer to green representing more use of contour features. We can see that on edges with rich texture or close to straight lines, the weights of the extracted texture features are greater, while on edges with single texture, the weights of the contour features are greater. (b) Examples of Textur/Contour ablation study. When the local contour shapes are similar, our network has difficulty matching them correctly using only contour features, and vice versa. However, these cases can be solved when the two features are fused.}}    
    \label{fig:weight and failure}
\end{figure}

\vspace{-0.2cm}
\paragraph{Performance on Generated Dataset.}
As shown in Table~\ref{Table of comparable method}, our network outperforms the comparison methods on both the full test set and on every set of difficulty levels. Jigsawnet shows a certain superiority in pair-searching on the high-difficulty set, probably because it treats the problem as a simple binary classification task; however, their pair-matching results are very unsatisfactory. 
It should be noted that we did not evaluate $\mathrm{Recall@1}$ and $\mathrm{NDCG@1}$ here due to the many-to-many pairings in our dataset, making it unreasonable to search only for the fragment with the highest similarity of searching. 
We can also observe that the performance of our proposed network shows reasonable variation with the difficulty of the data. This shows that our proposed network learns useful information and has more potential in complex situations. However, Jigsawnet and the rule-based method show relatively identical performance on test sets of different difficulty levels.
Furthermore, we also plot the distributions of HD, RE, and NTE for each method for a more intuitive comparison, as shown in Figure~\ref{fig:HD and RE}.

\vspace{-0.3cm}
\paragraph{Performance on Real Dataset.} 
\zhourixin{We also tested our method on the real dataset. In this experiment, both our method and the comparison method were trained on the generated dataset without retraining or fine-tuning. As shown in Table~\ref{Table of real dataset comparison} and Figure~\ref{fig:real matching}, PairingNet obtained satisfactory results for both pair-searching and -matching tasks. These results confirmed that Algorithm~\ref{alg:data generate} simulates real data well and our method has good generalization to real data.}


\paragraph{Visualization.} 
We show the top five search results of different methods for the examples of fragment pairs in Figure~\ref{fig:searching comparision}. Given a source fragment, our model can accurately identify the corresponding target fragment (marked with a green box). Furthermore, in situations where a source fragment can be paired with multiple target fragments, our method identifies all matching fragments within the top five search results. We also show examples of matching results for three difficulty levels in Figure~\ref{fig:comparision}.  Our method produces matching results that are consistent with the Ground Truth (GT),  while the comparison methods yield numerous incorrect results.

\yang{We also visualize the normalized self-gated weights to analyze their changes under different situations. As shown in Figure~\ref{fig:weight and failure} (a), we map the normalized weights to the edges of the fragments to show the influence of the texture/contour features. We classify the local parts of fragments into four types based on the information contained in texture and contour: those with rich texture and contour information (first quadrant), those with rich texture information (second quadrant), those with little texture and contour information (third quadrant), and those with rich contour information but with little texture information (fourth quadrant). The red color indicates that the weights of the texture features are relatively large, while the green color indicates that the contour feature weights are relatively large. The results show that our proposed network learned varying weights of the features for fragments with different characteristics.}

\vspace{-0.2cm}
\paragraph{Inference Time.} 
We test the inference time of different methods on the pair-searching task on our generated test set. Our network retrieves all fragment pairs in $73$ seconds, while the rule-based method takes $223$ hours, and Jigsawnet takes $10$ hours.

\begin{table}[t]
\centering
\caption{\yang{Results of ablation studies. Comparison of the impact of different feature fusion ways, feature modalities, and network layers employed in our proposed network on the performance of pair-searching and pair-matching tasks.}}
\label{Table of feature modality comparison and feature fusion comparison}
\resizebox{0.52\linewidth}{!}{ 
\begin{tabular}{cc|c|c|c|c|c}
\hline
\multicolumn{2}{c|}{}                                                                                          & Recall@5       & Recall@10      & NDCG@5         & NDCG@10        & RR             \\ \hline\hline
\multicolumn{1}{c|}{\multirow{3}{*}{\begin{tabular}[c]{@{}c@{}}Fusion \\ Way\end{tabular}}}       & Early fusion   & 0.448          & 0.559          & 0.376          & 0.413          & 0.658          \\ \cline{2-7}
\multicolumn{1}{c|}{}                                                                             & Concat         & 0.489          & 0.605          & \textbf{0.418} & \textbf{0.458} & 0.824          \\ \cline{2-7} 
\multicolumn{1}{c|}{}                                                                             & Weighted       & \textbf{0.493} & \textbf{0.608} & 0.417          & \textbf{0.458} & \textbf{0.835} \\ \hline\hline
\multicolumn{1}{c|}{\multirow{3}{*}{\begin{tabular}[c]{@{}c@{}}ResGCN \\  \end{tabular}}}              & 10           & -              & -              & \textbf{-}     & \textbf{-}     & 0.680 \\ \cline{2-7} 
\multicolumn{1}{c|}{}                                                                             & 14           & -              & -              & \textbf{-}     & \textbf{-}     & \textbf{0.835}            \\ \cline{2-7} 
\multicolumn{1}{c|}{}                                                                             & 18           & \textbf{-}     & \textbf{-}     & \textbf{-}     & \textbf{-}     & 0.759\\ \hline\hline
\multicolumn{1}{c|}{\multirow{3}{*}{\begin{tabular}[c]{@{}c@{}}Transformer\\ Encoder\end{tabular}}} & 3            & 0.488          & 0.600          & 0.410          & 0.450          & \textbf{-}                \\ \cline{2-7} 
\multicolumn{1}{c|}{}                                                                             & 4            & 0.488          & 0.606          & 0.411          & 0.452          & \textbf{-}                \\ \cline{2-7} 
\multicolumn{1}{c|}{}                                                                             & 5            & \textbf{0.493} & \textbf{0.608} & \textbf{0.417} & \textbf{0.458} & \textbf{-}                \\ \hline\hline
\multicolumn{1}{c|}{\multirow{3}{*}{\begin{tabular}[c]{@{}c@{}}Feature \\ Modality\end{tabular}}} & Contour-only    & 0.016          & 0.025          & 0.013          & 0.016          & 0.584          \\ \cline{2-7} 
\multicolumn{1}{c|}{}                                                                             & Texture-only    & 0.479          & 0.597          & 0.394          & 0.436          & 0.657          \\ \cline{2-7} 
\multicolumn{1}{c|}{}                                                                             & Contour + Texture & \textbf{0.493} & \textbf{0.608} & \textbf{0.417} & \textbf{0.458} & \textbf{0.835} \\ \hline
\end{tabular}

}
\end{table}

\vspace{-0.4cm}
\subsection{Ablation Study}
 \vspace{-0.2cm}
 
\paragraph{Network Structures.}
\yang{We explore the performance of our network under different structures, as shown in Table~\ref{Table of feature modality comparison and feature fusion comparison}. First, we test three feature fusion ways. In early fusion, instead of using two separate GCNs for feature extraction, we concatenate the two modalities and use only one GCN to learn the features. We also test using a simple concatenation to replace the self-gated fusion module. The results show that the performance of both pair-searching and -matching decreases. Additionally, we explore the impact of different numbers of layers on our network.}




\vspace{-0.3cm}
\paragraph{Texture/Contour.}
\yang{To study the impact of contour and texture features on the performance of our network, we conducted experiments using only one modality (either contour or texture) and compared the final results. As shown in Table~\ref{Table of feature modality comparison and feature fusion comparison}, the comparison results indicate that both modalities perform slightly worse when used independently. When both modalities are used in conjunction, our network achieves the best performance. We also show examples of failed pair-matching when the two modalities are used individually in Figure~\ref{fig:weight and failure} (b).

Besides, as shown in Figure~\ref{fig:weight and failure} (a), it can be seen numerically that our network tends to assign more weight to the texture features. This is reasonable because textures usually contain richer information than the contours of image fragments, enabling the network to learn more discriminative features.}

\vspace{-0.3cm}
\paragraph{Others.} We report the results of ablation studies on patch size, contour encoding, and hyperparameters in our supplementary material.

%% file: sections/6_conclusion.tex
\vspace{-0.2cm}
\section{Conclusions}
\vspace{-0.4cm}
In this paper, to tackle the image fragment pair-searching and -matching tasks, we propose a novel network, PairingNet, to deduce the adjacent segments of fragments and encode the global features of each fragment. In PairingNet, ResGCN is employed as the backbone to extract the local contour and texture features of fragments. For pair-searching, a linear transformer-based module processes these local features leveraging the contrast of paired fragments and unpaired fragments. For pair-matching, we design a weighted fusion module to dynamically fuse the extracted local contour and texture features. Furthermore, to provide a foundation for learning-based methods, we design a cutting algorithm to generate a dataset by simulating real image fragments. Comprehensive experiments are conducted on our generated dataset and a real dataset, the results of which demonstrate the effectiveness of our PairingNet compared to existing methods. In future work, we will explore how to extend our work to other general technique, e.g. puzzles inspired MAE~\cite{he2022masked}, JiGen~\cite{carlucci2019domain} for self-supervised learning and domain generalization. The fragment generation algorithm that we designed can also create more diverse samples for image inpainting tasks~\cite{liu2018image, li2022mat}.

\zhourixinTwo{
\noindent\textbf{Limitations} While our research has made significant strides, fragment restoration remains a challenging task. Our dataset provides superior edge matching features. The task of fragment restoration becomes particularly arduous when the edges of the fragments are smooth and the texture is monotonous (for instance, in the case of shredded white paper), or when there is a substantial gap in the damaged edges.


\noindent\textbf{Acknowledgements} This work was supported by the Young Scientists Fund of the National Natural Science Foundation of China (Grant No.62206106). We also feel grateful to Yifan Jin for his exploration in the preliminary work.
}



%% file: sections/7_supp.tex
\definecolor{myblue}{RGB}{47,73,156} 
\definecolor{myorange}{RGB}{233,112,57} 
\definecolor{mygreen}{RGB}{2,125,63} 

\fancypagestyle{firstpagefooter}{
	\fancyhf{} 
	\renewcommand{\headrulewidth}{0pt} 
	\renewcommand{\footrulewidth}{1pt} 
	\fancyfoot[L]{\tiny *Corresponding authors \\
 $\dagger$ Engineering Research Center of Knowledge-Driven Human-Machine Intelligence, MoE, China \\
 $\ddagger$ Key Laboratory of Ancient Chinese Script, Culture Relics and Artificial Intelligence, Jilin Province} 
}


\title{PairingNet: A Learning-based Pair-searching and -matching Network for Image Fragments \textmd{Supplementary Materials}} 

\titlerunning{PairingNet}

\author{Rixin Zhou\inst{1}\orcidlink{0009-0005-2670-609X} \and
Ding Xia\inst{2}\orcidlink{0000-0002-4800-1112} \and
Yi Zhang\inst{3}\orcidlink{0000-0003-0294-8973} \and
Honglin Pang\inst{1}\orcidlink{0009-0005-4870-6605} \and
Xi Yang\inst{1,}\textsuperscript{*,$\dagger$, $\ddagger$}\orcidlink{0000-0001-5039-3680} \and
Chuntao Li\inst{4,}\textsuperscript{*, $\ddagger$}\orcidlink{0000-0001-9836-1493}}

\authorrunning{Zhou et al.}

\institute{School of Artificial Intelligence, Jilin University \and
The University of Tokyo \and
National University of Defense Technology, College of Electrionic Engineering\and 
School of Archaeology, Jilin University \\
\email{\{zhourx22\}@mails.jlu.edu.cn}
\email{\{lct33\}@jlu.edu.cn} \email{\{dingxia1995,panghlwork,earthyangxi\}@gmail.com}
\email{\{2356711993\}@qq.com}
}

\maketitle

\pagestyle{headings}
\thispagestyle{firstpagefooter}

\section{Overview}

The supplementary materials that we provide are organized as follows:


\begin{itemize}
\setlength{\itemsep}{0pt}
\setlength{\parsep}{0pt}
\setlength{\parskip}{0pt}



\item Supplementary Material.pdf (this file) 
\item Dataset and Sourcecode (\href{https://github.com/zhourixin/PairingNet}{link})
  \begin{itemize}
  \setlength{\itemsep}{0pt}
  \setlength{\parsep}{0pt}
  \setlength{\parskip}{0pt}
    \item[*] Scanned real dataset
    \item[*] Our generated dataset
    \item[*] Code of dataset generation 
    \item[*] Code of our proposed PairingNet
    \item[*] Code of comparison methods (Jigsawnet and rule-based methods)
  \end{itemize}
  
\end{itemize}

\section{Generated Dataset Examples} 
\begin{figure*}[t]
    \includegraphics[width=1\linewidth]{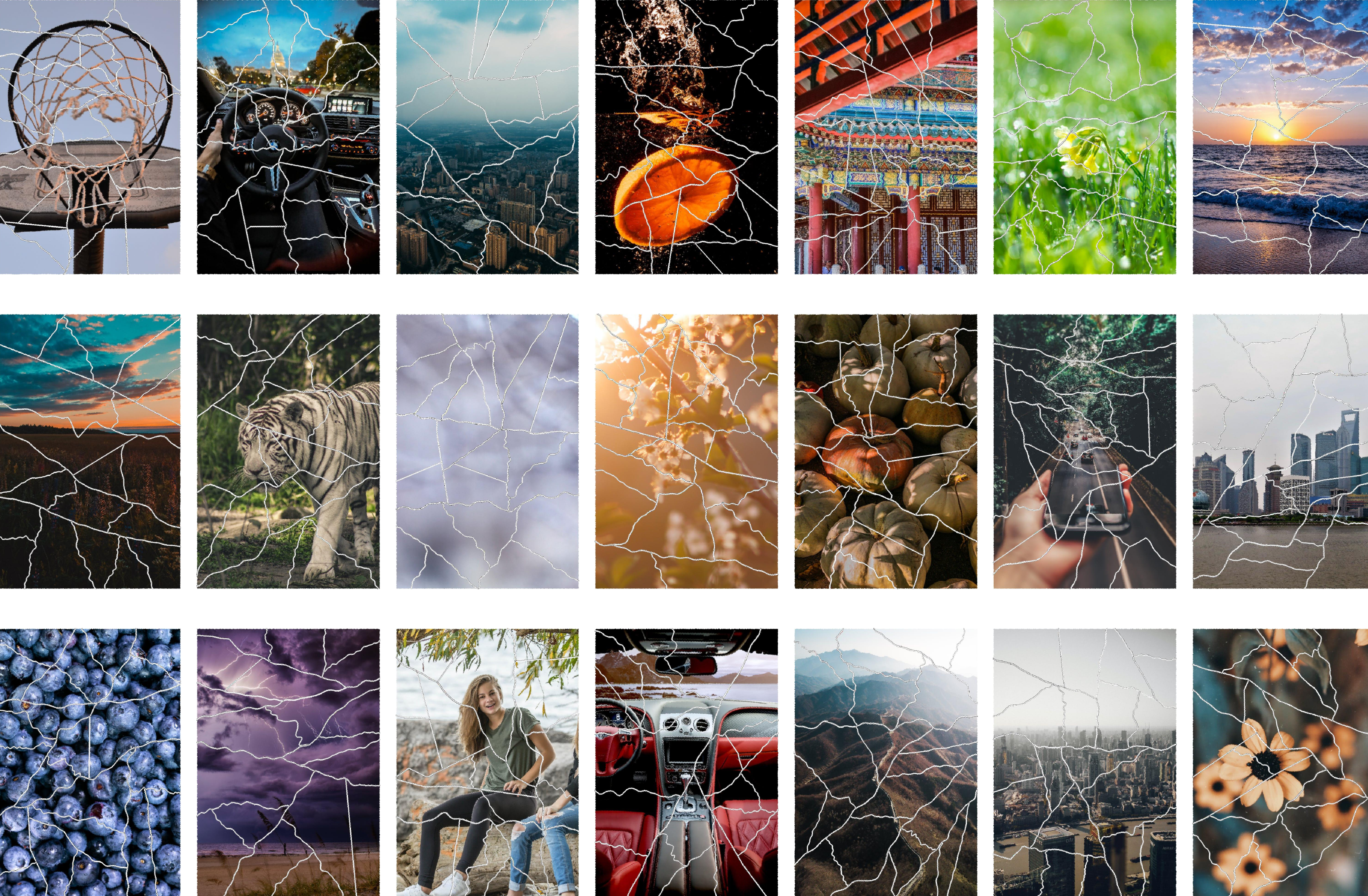}
    \caption{Cutting examples in our generated dataset, which are rich in shapes and patterns.}
    \label{fig:multi image}
\end{figure*}

The cutting algorithm we designed is implemented in Python and takes an average of $11$ seconds to tear a complete image into the required fragments. We show more cutting results generated in our dataset, as shown in Figure~\ref{fig:multi image}. The fragments we generate are rich in shapes and patterns and have a similar effect to shredding by hand.

\begin{figure*}[t]
\centering
    \includegraphics[width=1\linewidth]{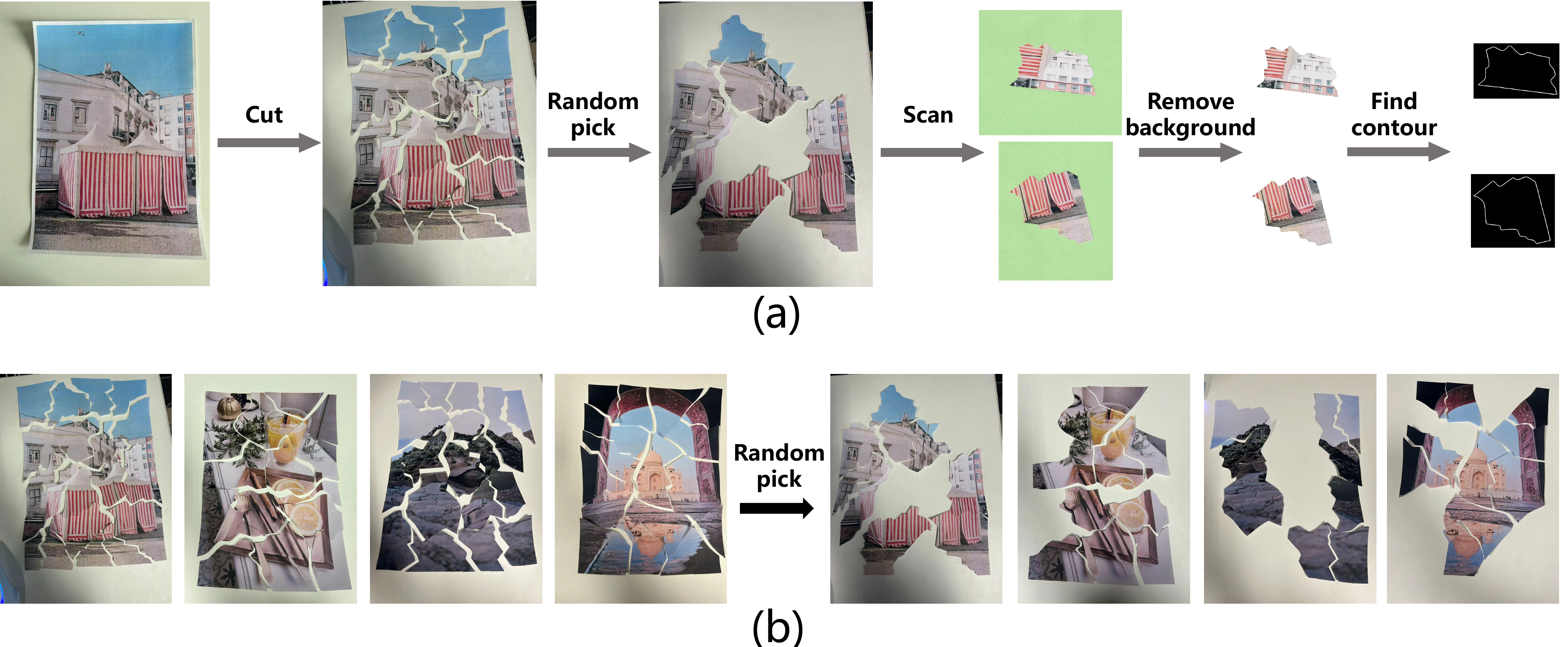}
    \caption{(a) The process of generating real dataset. (b) More examples of processing a complete real image into multiple fragments.}
    \label{fig:make real dataset}
\end{figure*}

\begin{table}[!ht]
    \centering
    \caption{Statistical information of the real dataset.}
    \resizebox{0.55\linewidth}{!}{ 
    \begin{tabular}{c|c|c|c}
    \hline
        Complete Image &  Image Resolution &  Fragments & Pairs \\ \hline \hline
        34 & 2480x3508 & 320 & 202 \\ \hline
    \end{tabular}
    }
    \label{table: real dataset satistic}
\end{table}

\section{Real Dataset Details} 

\zhourixin{In this section, we detail the methodology employed for the creation of our real dataset. The process begins with the printing of 34 complete images, which are subsequently excluded from our generated dataset. A random assortment of these fragments is discarded. To simulate real-world situations, we randomly discard some fragments. The remaining fragments are then digitized using a laser scanner. These digital images undergo post-processing steps, which include the removal of the background and extraction of contours. The Scale-Invariant Feature Transform (SIFT) algorithm is utilized to identify key points and descriptors of both the complete image and the fragment image. The Fast Library for Approximate Nearest Neighbors (FLANN) matcher is used for descriptor matching, which allows us to annotate the matching points between fragment pairs. 

Figure~\ref{fig:make real dataset} (b) provides additional examples of the transformation of a complete real image into multiple fragments. On average, the processing of a complete real image takes approximately one hour. The entire process was executed by two researchers over a span of 17 hours.  The relevant statistical information of the dataset is presented in Table~\ref{table: real dataset satistic}
}



\begin{figure}[t]
    \centering
    \includegraphics[width=1\textwidth]{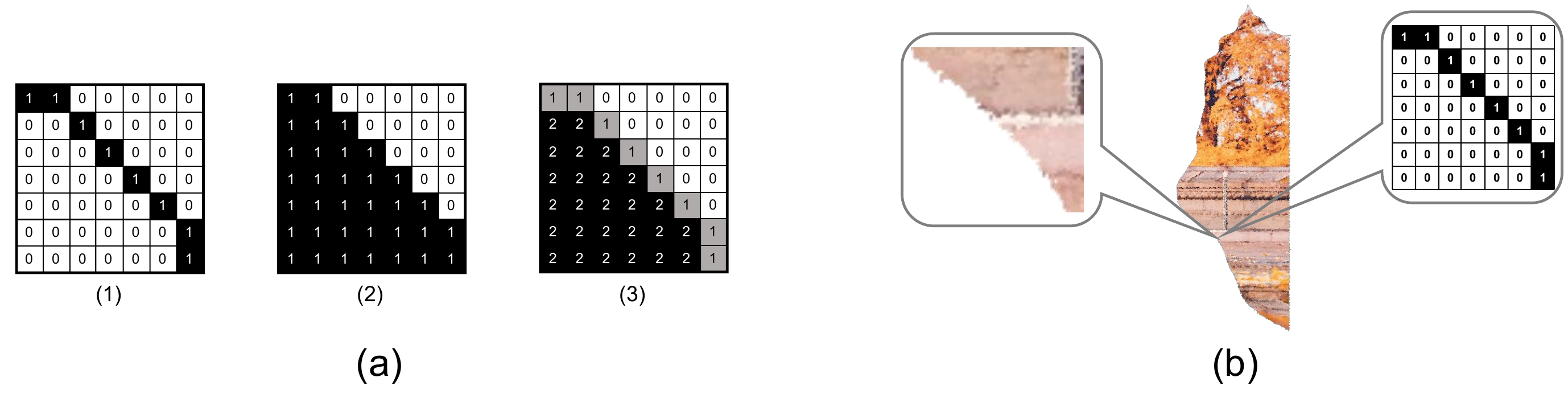}
    \caption{Coutour encodings. (a) Three contour encodings: (1) Edge-only, (2) Inside-Outside, (3) Edge + Inside-Outside. (b) Example of Edge-only encoding we used.}
    \label{fig:contour encoding methods}
\end{figure}

\section{Contour Encoding Details}

We design three contour encodings as shown in Figure~\ref{fig:contour encoding methods}: (1) Edge-only; (2) Inside-Outside; (3) Edge + Inside-Outside. Encoding (1) describes where the contour is located. Encoding (2) describes which positions are fragmented parts and which are not. Encoding (3) combines the above two encodings. We also test the impact of different patch sizes on performance in the ablation studies.

\begin{table*}[t]
\centering
\caption{Detailed results of ablation studies.}
\label{Table of all ablation study}
\resizebox{1\linewidth}{!}{ 
\begin{tabular}{ccc|c|c|c|c|c|c|c|c|c|c}
\hline
\multicolumn{3}{c|}{}                                                                                                                                                   & Recall@5       & Recall@10      & Recall@20      & NDCG@5         & NDCG@10        & NDCG@20        & RR $\uparrow$     & HD $\downarrow$     & RE $\downarrow$   & NTE $\downarrow$       \\ \hline\hline

\multicolumn{2}{c|}{\multirow{3}{*}{\begin{tabular}[c]{@{}c@{}}Contour\\ Encoding\end{tabular}}}                                                    & (1)                & -              & -              & -              & -     & -     & -              & \textbf{0.835} & \textbf{43.180}   & \textbf{0.352} & \textbf{7.735$\times10^{-4}$}  \\ \cline{3-13} 
\multicolumn{2}{c|}{}                                                                                                                              & (2)                & -              & -              & -              & -     & -     & -              & 0.695          & 75.423           & 0.576          & 13.716$\times10^{-4}$          \\ \cline{3-13} 
\multicolumn{2}{c|}{}                                                                                                                              & (3)                & -     & -     & -              & -     & -     & -              & 0.671          & 84.982           & 0.641          & 15.078$\times10^{-4}$          \\\hline\hline
\multicolumn{2}{c|}{\multirow{3}{*}{\begin{tabular}[c]{@{}c@{}}Patch\\ Size\end{tabular}}}                                                         & 3$\times$3                & -              & -              & -              & -     & -     & -              & 0.714          & 72.477           & 0.566          & 13.291$\times10^{-4}$          \\ \cline{3-13} 
\multicolumn{2}{c|}{}                                                                                                                              & 7$\times$7                & -              & -              & -              & -     & -     & -              & \textbf{0.835} & \textbf{43.180}   & \textbf{0.352} & \textbf{7.735$\times10^{-4}$}  \\ \cline{3-13} 
\multicolumn{2}{c|}{}                                                                                                                              & 11$\times$11              & -     & -     & -              & -     & -     & -              & 0.663          & 84.751           & 0.649          & 15.306$\times10^{-4}$          \\ \hline\hline
\multicolumn{2}{c|}{\multirow{3}{*}{ResGCN Layer}}                                                                                                 & 10                 & -     & -     & -              & -     & -     & -              & 0.68           & 81.163           & 0.610          & 14.847$\times10^{-4}$          \\ \cline{3-13} 
\multicolumn{2}{c|}{}                                                                                                                              & 14                 & -     & -     & -              & -     & -     & -              & \textbf{0.835} & \textbf{43.180}   & \textbf{0.352} & \textbf{7.735$\times10^{-4}$}  \\ \cline{3-13} 
\multicolumn{2}{c|}{}                                                                                                                              & 18                 & -     & -     & -              & -     & -     & -              & 0.759          & 62.51            & 0.484          & 11.165$\times10^{-4}$          \\  \hline\hline
\multicolumn{2}{c|}{\multirow{3}{*}{\begin{tabular}[c]{@{}c@{}}Transformer\\ Encoder Layer\end{tabular}}}                                          & 3                  & 0.488          & 0.600          & 0.697          & 0.410          & 0.450          & 0.476          & -     & -       & -     & -          \\ \cline{3-13} 
\multicolumn{2}{c|}{}                                                                                                                              & 4                  & 0.488          & 0.606          & 0.711          & 0.411          & 0.452          & 0.481          & -     & -       & -     & -          \\ \cline{3-13} 
\multicolumn{2}{c|}{}                                                                                                                              & 5                  & \textbf{0.493} & \textbf{0.608} & \textbf{0.714} & \textbf{0.417} & \textbf{0.458} & \textbf{0.487} & -     & -       & -     & -          \\  \hline\hline
\multicolumn{2}{c|}{\multirow{3}{*}{\begin{tabular}[c]{@{}c@{}}Fusion\\ Way\end{tabular}}}                                                         & Early fusion       & 0.448          & 0.559          & 0.672          & 0.376          & 0.413          & 0.446          & 0.658          & 85.558           & 0.636          & 15.547$\times10^{-4}$          \\ \cline{3-13} 
\multicolumn{2}{c|}{}                                                                                                                              & Concat             & 0.489          & 0.605          & 0.706          & \textbf{0.418} & \textbf{0.458} & 0.486          & 0.824          & 45.521           & 0.359          & 8.269$\times10^{-4}$           \\ \cline{3-13} 
\multicolumn{2}{c|}{}                                                                                                                              & Weighted           & \textbf{0.493} & \textbf{0.608} & \textbf{0.714} & 0.417          & \textbf{0.458} & \textbf{0.487} & \textbf{0.835} & \textbf{43.180}  & \textbf{0.352} & \textbf{7.735$\times10^{-4}$}  \\ \hline\hline
\multicolumn{2}{c|}{\multirow{2}{*}{\begin{tabular}[c]{@{}c@{}}Feature Fusion \\ in Pair-searching Module\end{tabular}}}                           & Element-wise Addition      & 0.489          & 0.605          & 0.706          & \textbf{0.418} & \textbf{0.459} & 0.486          & -               & -                 & -               &   -                  \\ \cline{3-13} 
\multicolumn{2}{c|}{}                                                                                                                              & Channel-wise Concatenation & \textbf{0.493} & \textbf{0.608} & \textbf{0.714} & 0.417          & 0.458          & \textbf{0.487} &  -              &    -              &   -             &    -                 \\ \hline\hline
\multicolumn{2}{c|}{\multirow{3}{*}{\begin{tabular}[c]{@{}c@{}}Focal Loss\\ Paramaters\end{tabular}}}                                               & ${\beta}_1$=0.25, $\gamma$=4 & -     & -     & -              & -     & -     & -              & 0.514          & 124.88           & 0.886          & 21.972$\times10^{-4}$          \\ \cline{3-13} 
\multicolumn{2}{c|}{}                                                                                                                              & ${\beta}_1$=0.4, $\gamma$=6  & -     & -     & -              & -     & -     & -              & 0.752          & 65.181           & 0.505          & 11.631$\times10^{-4}$          \\ \cline{3-13} 
\multicolumn{2}{c|}{}                                                                                                                              & ${\beta}_1$=0.55, $\gamma$=8 & -     & -     & -              & -     & -     & -              & \textbf{0.835} & \textbf{43.180}   & \textbf{0.352} & \textbf{7.735$\times10^{-4}$}  \\ \hline\hline
\multicolumn{2}{c|}{\multirow{3}{*}{\begin{tabular}[c]{@{}c@{}}InfoNCE\\ Temperature\end{tabular}}}                                               & 0.07               & 0.485          & 0.593          & 0.692          & 0.413          & 0.451          & 0.478          & -     & -       & -     & -          \\ \cline{3-13} 
\multicolumn{2}{c|}{}                                                                                                                              & 0.12               & \textbf{0.493} & \textbf{0.608} & \textbf{0.714} & \textbf{0.417} & \textbf{0.458} & \textbf{0.487} & -     & -       & -     & -          \\ \cline{3-13} 
\multicolumn{2}{c|}{}                                                                                                                              & 0.2                & 0.474          & 0.579          & 0.679          & 0.394          & 0.430          & 0.457          & -     & -       & -     & -          \\ \hline\hline
\multicolumn{1}{c|}{\multirow{9}{*}{\begin{tabular}[c]{@{}c@{}}Feature\\ Modality\end{tabular}}} & \multicolumn{1}{c|}{\multirow{3}{*}{RGB}}       & Contour-only       & 0.016          & 0.025          & 0.049          & 0.013          & 0.016          & 0.023          & 0.584          & 122.953          & 0.846          & 19.806$\times10^{-4}$          \\ \cline{3-13} 
\multicolumn{1}{c|}{}                                                                            & \multicolumn{1}{c|}{}                           & Texture-only       & 0.479          & 0.597          & 0.705          & 0.394          & 0.436          & 0.466          & 0.657          & 85.438           & 0.684          & 15.872$\times10^{-4}$          \\ \cline{3-13} 
\multicolumn{1}{c|}{}                                                                            & \multicolumn{1}{c|}{}                           & Contour+Texture    & \textbf{0.493} & \textbf{0.608} & \textbf{0.714} & \textbf{0.417} & \textbf{0.458} & \textbf{0.487} & \textbf{0.835} & \textbf{43.180}  & \textbf{0.352} & \textbf{7.735$\times10^{-4}$}  \\ \cline{2-13} 
\multicolumn{1}{c|}{}                                                                            & \multicolumn{1}{c|}{\multirow{3}{*}{Grayscale}} & Contour-only       & 0.018          & 0.032          & 0.053          & 0.013          & 0.019          & 0.025          & 0.535          & 138.238          & 0.942          & 22.314$\times10^{-4}$          \\ \cline{3-13} 
\multicolumn{1}{c|}{}                                                                            & \multicolumn{1}{c|}{}                           & Texture-only       & 0.258          & 0.355          & 0.470          & 0.205          & 0.24           & 0.271          & 0.633          & 94.706           & 0.700          & 17.147$\times10^{-4}$          \\ \cline{3-13} 
\multicolumn{1}{c|}{}                                                                            & \multicolumn{1}{c|}{}                           & Contour+Texture    & \textbf{0.281} & \textbf{0.379} & \textbf{0.487} & \textbf{0.226} & \textbf{0.261} & \textbf{0.291} & \textbf{0.833} & \textbf{44.917}  & \textbf{0.365} & \textbf{7.738$\times10^{-4}$}  \\ \cline{2-13} 
\multicolumn{1}{c|}{}                                                                            & \multicolumn{1}{c|}{\multirow{3}{*}{Binarized}} & Contour-only       & \textbf{0.086} & \textbf{0.124} & \textbf{0.184} & \textbf{0.069} & \textbf{0.083} & \textbf{0.100} & \textbf{0.562} & \textbf{126.609} & \textbf{0.857} & \textbf{20.459$\times10^{-4}$} \\ \cline{3-13} 
\multicolumn{1}{c|}{}                                                                            & \multicolumn{1}{c|}{}                           & Texture-only       & 0.027          & 0.048          & 0.082          & 0.019          & 0.027          & 0.037          & 0.000              & 232.263          & 1.562          & 42.034$\times10^{-4}$          \\ \cline{3-13} 
\multicolumn{1}{c|}{}                                                                            & \multicolumn{1}{c|}{}                           & Contour+Texture    & 0.036          & 0.051          & 0.083          & 0.029          & 0.034          & 0.043          & 0.553          & 127.997          & 0.888          & 21.072$\times10^{-4}$          \\ \hline\hline

\multicolumn{3}{c|}{Use Graph U-Nets as Backbone}                                                                                                                        & -     & -     & -     & -     & -     & -     &    0.629   &97.111                  &0.744                &17.001$\times10^{-4}$                     \\
\hline\hline
\multicolumn{3}{c|}{\begin{tabular}[c]{@{}c@{}}Train Pair-searching module using Pair-matching Features\end{tabular}}                               & 0.434          & 0.535          & 0.632          & 0.358          & 0.392          & 0.419          & -     & -       & -     & -          \\ 

\hline

\end{tabular}
}
\end{table*}

\section{Ablation Study}
\label{Ablation study}
We provide detailed ablation study results as shown in Table~\ref{Table of all ablation study}.

\subsection{Network Details}

\paragraph{Contour Encoding Methods.}
We tested the impact of three contour encodings on the pair-matching task. Experimental results show that Encoding (1) Egde-only achieves the best matching performance.

\vspace{-0.3cm}
\paragraph{Patch Size.}
We use a patch size of $7\times7$ to encode contours in our proposed network. We also tested the impact of patch sizes $3\times3$ and $11\times11$ on the pair-matching task: however, they did not achieve better performance.

\vspace{-0.3cm}
\paragraph{ResGCN Layer.}
The ResGCN encoder in both branches of our network contains $14$ GCN layers, with each node connecting $8$ nodes on both sides of its vicinity. The length of the contour points is unified to the maximum length $2900$, and zero-padding is used. We tested the impact of the $10$ and $18$ layers on the pair-matching task. When using more or fewer GCN layers, the pair-matching performance of our model decreases. This shows that simply increasing the capacity of the encoder does not always lead to better performance.

\vspace{-0.3cm}
\paragraph{Transformer Encoder Layer.}
The linear transformer encoder in our network contains $5$ layers, and the input length is truncated to $1408$. We tested the impact of $3$ and $4$ layers on pair-searching performance. When using more layers of Transformer Encoder, the network has better searching performance. Due to computational resource constraints, we did not further test higher capacities. 

\vspace{-0.3cm}
\paragraph{Fusion way.} 
In addition to the shortened table, we also provide all results of the three feature fusion methods to show the impact on our proposed network.

\vspace{-0.3cm}
\paragraph{Feature Fusion in Pair-searching Module.}
Due to better performance, we replaced the element-wise addition operation of the feature fusion in the pair-searching module with channel-wise concatenation.


\vspace{-0.3cm}
\paragraph{Parameters in Loss Functions.}
For the two-step training strategy, we first train the feature extraction network and the pair-matching module by our pair-matching loss. Freeze them and then train the pair-searching module by our pair-searching loss.
In the Focal loss, the balancing weights ${\beta}_1$ was set to $0.55$ and the decay factor $\gamma$ was set to $8$. We also tested two other sets of parameter combinations, ${\beta}_1=0.25$, $\gamma=4$ and $\beta_1=0.4$, $\gamma=6$, respectively. In the  InfoNCE loss, we set the temperature to $0.12$. We also tested the impact of setting the temperature to $0.07$ and $0.2$. We chose the parameters that achieved optimal performance.

\subsection{Reduce Texture Information}

\begin{figure}[t]
\centering
    \includegraphics[width=1\textwidth]{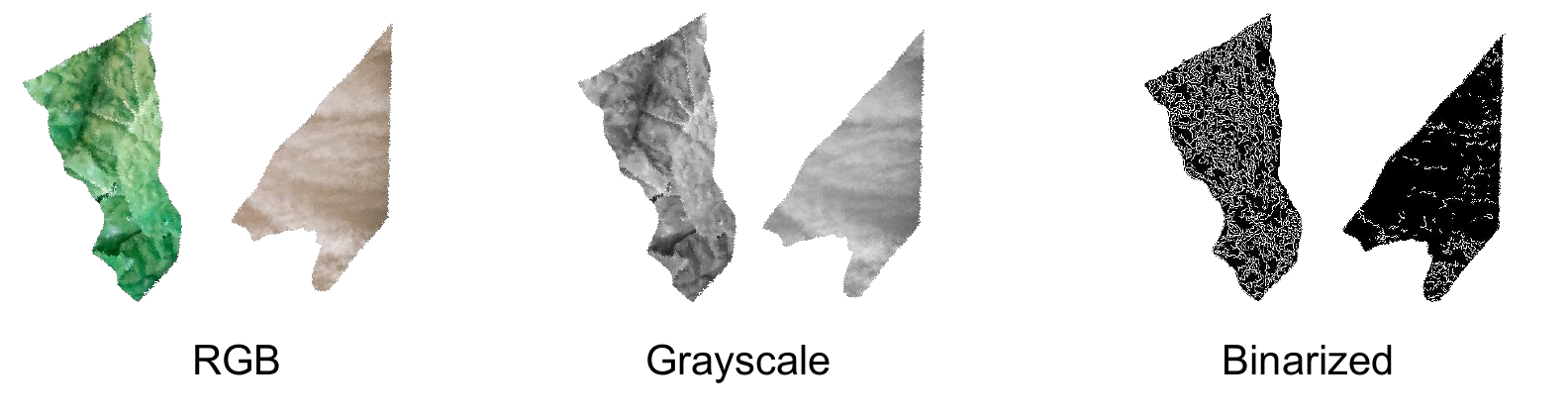}
    \caption{Examples of the RGB, grayscale, and binarized fragments.}
    \label{fig:binarize and gray}
\end{figure}

In the ablation study of feature modality, we can see that texture features contribute more performance compared to contour features. To further verify the necessity of extracting contour features, we processed the original RGB fragments into grayscale and binarized them using the Canny edge detector (Figure~\ref{fig:binarize and gray}) to reduce the texture information and show the impact on our tasks.

The evaluation results got worse, since the texture information is lost in grayscale fragments and binarized ones have almost no texture features. Then, when we input binarized fragments, the contour-only modality achieved the best performance, showing the effectiveness of contour features.


\subsection{Others}
\paragraph{Backbone.}
We tested another commonly used graph-based network, Graph U-Nets~\cite{gao2019graph}, as our backbone; however, the performance decreased.

\vspace{-0.3cm}
\paragraph{Train Pair-searching module using Pair-matching Features.}
We also tested using the features fused by the pair-matching module to train the pair-searching module; however, the performance declined. This shows that the features required for pair-searching and -matching tasks are not exactly the same.


\subsection{Visualization}

\begin{figure}[t]
\centering
    \includegraphics[width=1\linewidth]{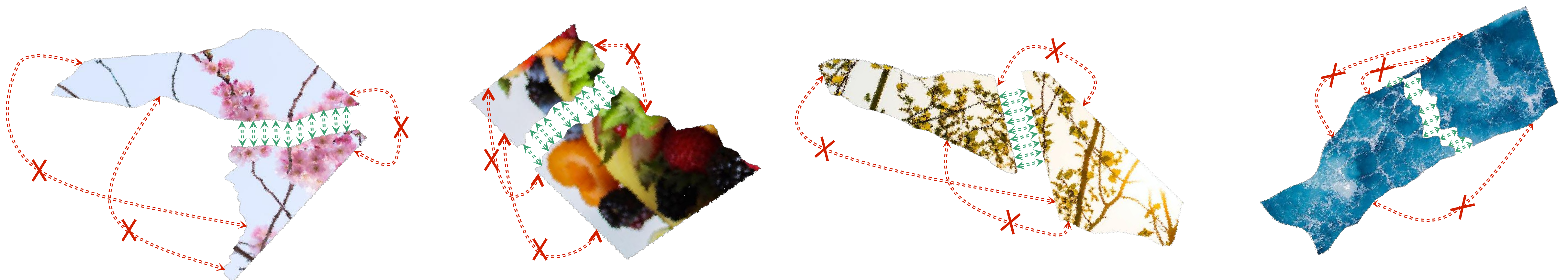}
    \caption{Pair-matching correspondences. Lines show the correspondences predicted by our network between fragment pairs. Green is used to indicate correctly paired points, while red is used to indicate incorrectly paired points.}
    \label{fig:corresponding}
\end{figure}

\begin{figure}[t]
\centering
    \includegraphics[width=0.6\linewidth]{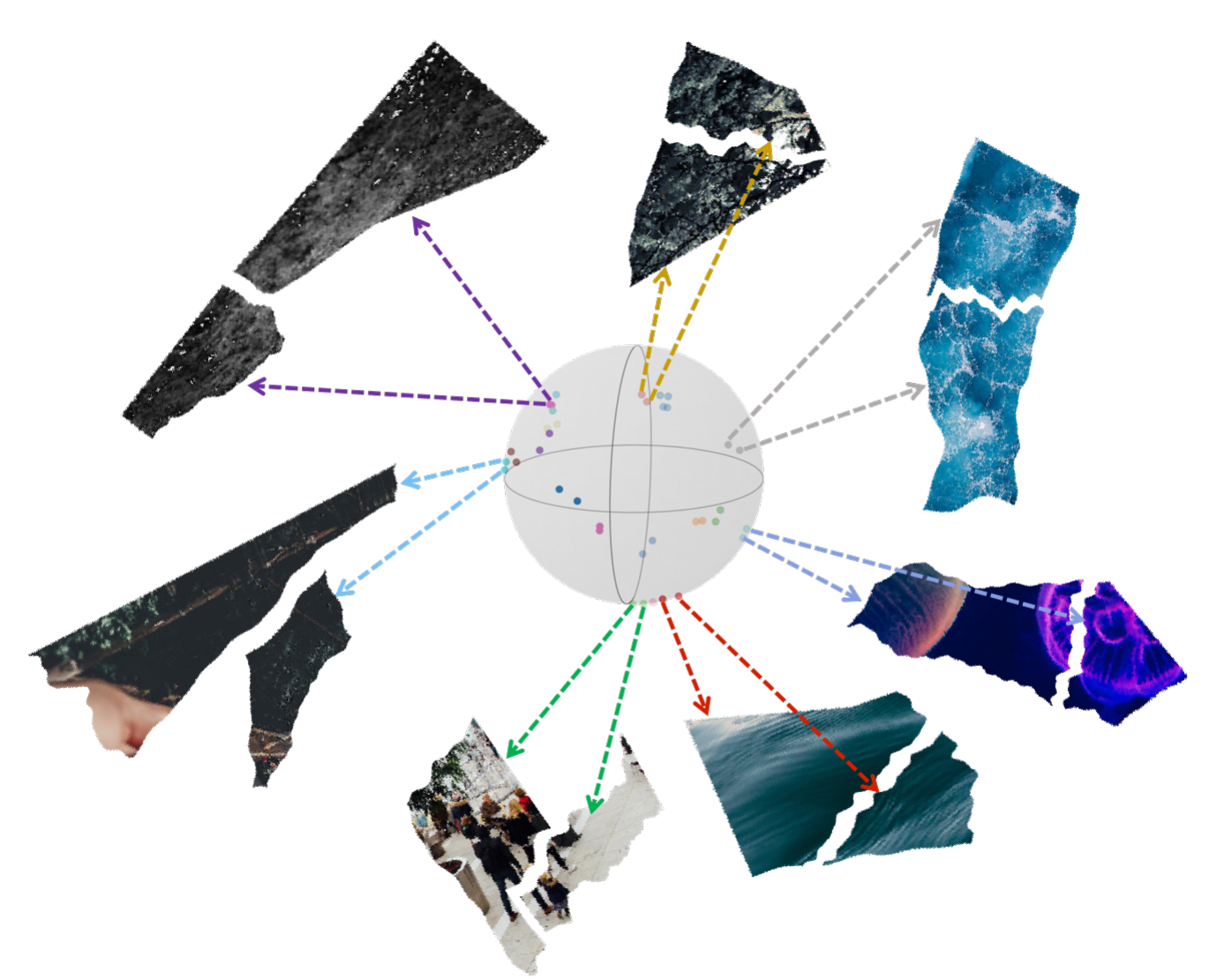}
    \caption{Visualization of the features extracted by our pair-searching module. The paired fragments are represented by the same color.}
    \label{fig:sphere}
\end{figure}

\begin{figure}[t]
\centering
    \includegraphics[width=0.6\linewidth]{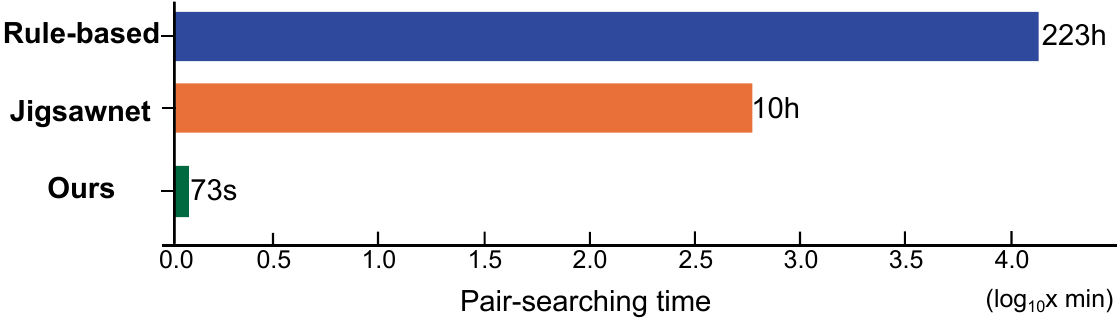}
    \caption{Total inference time for different methods on the pair-searching task. This comparison is numerically unfair because the experiments were performed on different devices; however, it is useful in practice.}
    \label{fig:time}
\end{figure}

\begin{figure}[t]
\centering
    \includegraphics[width=1\linewidth]{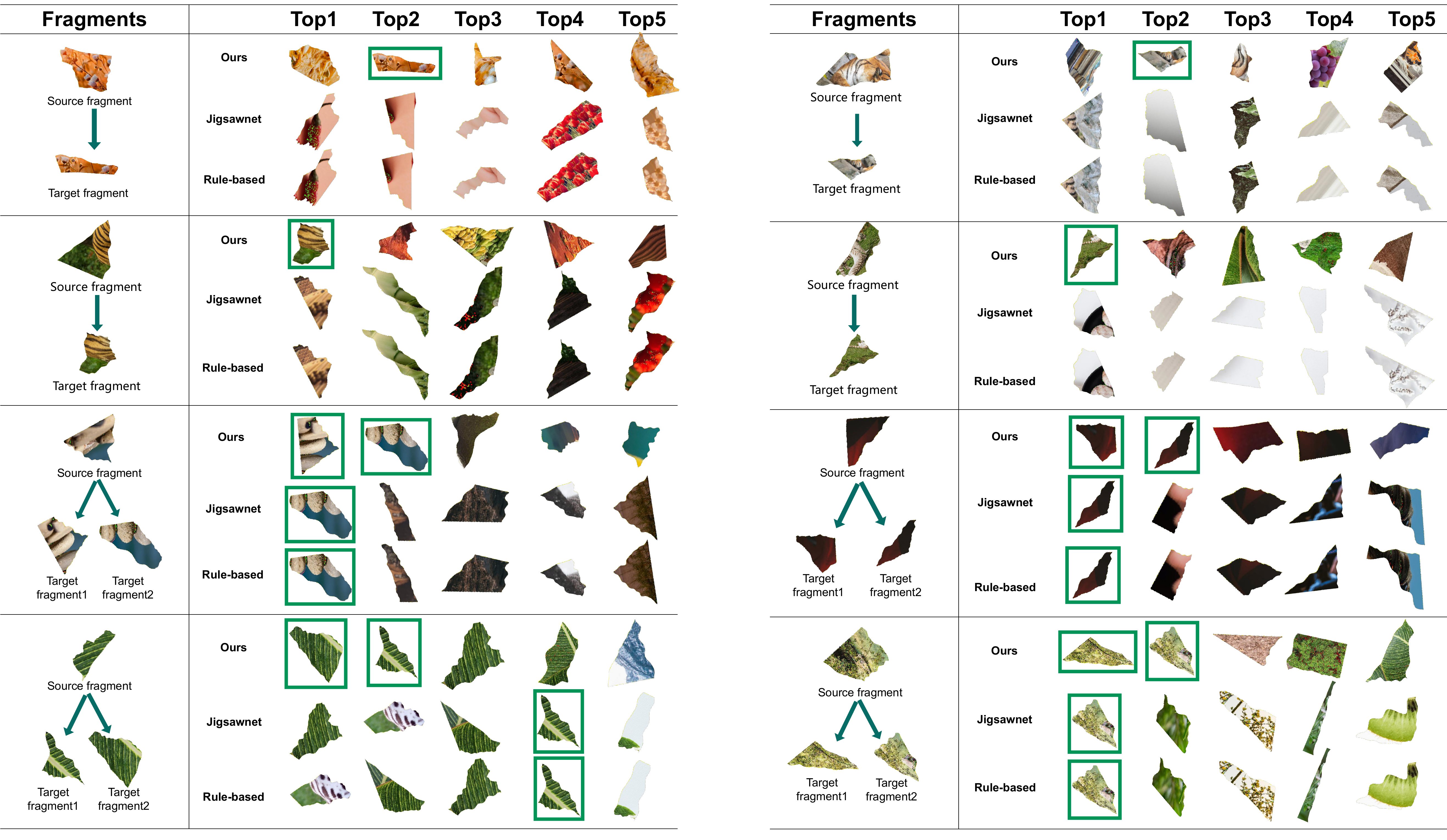}
    \caption{Examples of pair-searching results. Our network is able to accurately identify the corresponding target fragment (highlighted with a green box).}
    \label{fig:more searching}
\end{figure}

\begin{figure}[h]
\centering
    \includegraphics[width=1\linewidth]{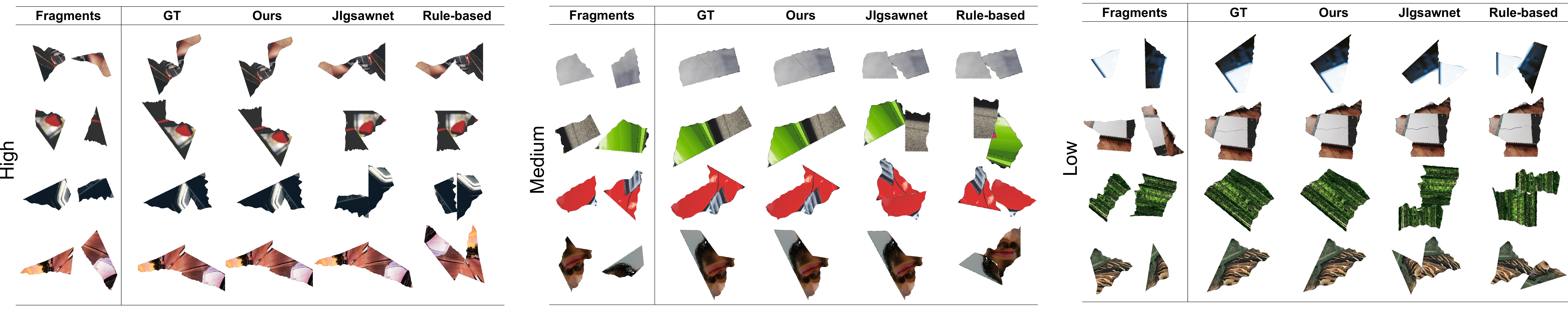}
    \caption{Examples of pair-matching results for three difficulty levels. Our proposed network achieves satisfactory matching results.}
    \label{fig:more matching}
\end{figure}

\begin{figure}[h]
\centering
    \includegraphics[width=1\linewidth]{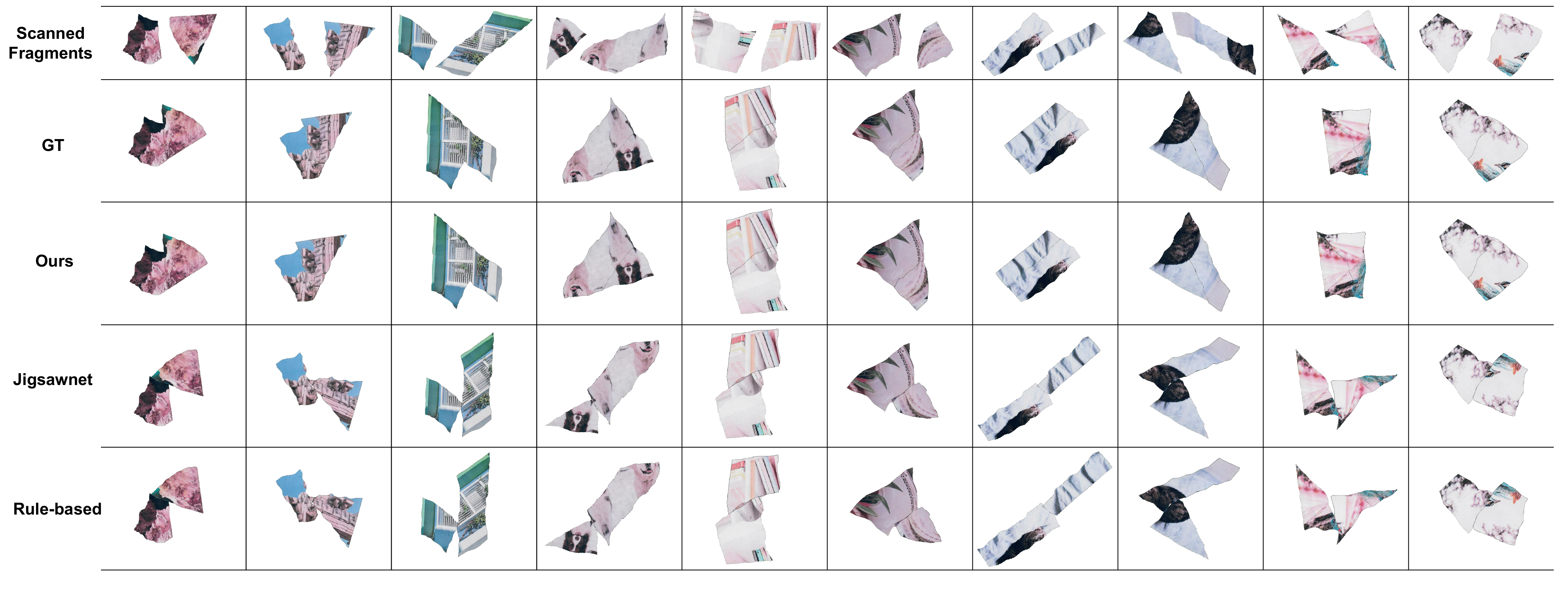}
    \caption{Examples of pair-matching results of PairingNet on real dataset. Without retraining, PairingNet achieves satisfactory matching results on real fragments.}
    \label{fig:more matching}
\end{figure}

We visualize pair-matching correspondences between the fragments predicted by our network, as shown in Figure~\ref{fig:corresponding}. The correct matching points are connected by green lines, while incorrectly predicted matching points are connected by red lines. For any two fragments, in addition to the optimal matching points, there are also suboptimal matching points that may interfere with the network performance. For example, the locations connected by the red lines in Figure~\ref{fig:corresponding} also exhibit matchable local contours or textures. Only a sufficiently high-performance network can predict as many correct matching points as possible and minimize the interference of sub-optimal matching points on the results.

We perform dimensionality reduction visualization on the features extracted by our network to show the effectiveness of contrastive loss in the pair-searching task. Compared with Principal Component Analysis (PCA), which prefers to maintain the global structure of feature space, and t-SNE, which focuses on maintaining the local structure, we select PaCMAP~\cite{wang2021understanding}, a dimensionality reduction technique that preserves both global and local structures of the feature space,  to reduce the features extracted by the pair-searching module to three dimensions. Subsequently, we normalize the features of each fragment and map them onto the surface of the unit sphere. We show examples in Figure~\ref{fig:sphere}. We can see that the features of fragment pairs represented by the same color points are close to each other, while the features of different pairs are pushed further apart.

\subsection{More Comparison Results}

We present more results of our network and comparison methods on pair-searching and -matching tasks, as shown in Figures~\ref{fig:more searching} and \ref{fig:more matching}. It can be intuitively seen that in the pair-matching task, the results obtained by our method are closer to GT, while the comparison method produces many inaccurate results. In the pair-searching task, whether it is one-to-one matching or one-to-many matching, our method can find the target fragments in the Top 5 search results. The comparison method can only find a part of the matching target fragments, and sometimes it cannot even find correct target fragments in the Top 5 search results.  In the pair-searching task, our network finds the paired fragments in the Top 5 search results, whether it is one-to-one matching or one-to-many matching. However, the comparison methods can only find a part of the matching target fragments, and even cannot find correct target fragments in the Top 5. 
In the pair-matching task, the results obtained by our network are closer to GT, while the comparison methods produce inaccurate results.

\subsection{Inference time}
We test the inference time of different methods in the pair-searching task on our generated test set, as shown in Figure~\ref{fig:time}. For ease of presentation, we have plotted the logarithm of the inference time for the pair-searching task. \zhourixin{In the submitted manuscript, we highlight that the comparison methods are specifically tailored for computations accelerated by CPUs. These methods were tested on a robust CPU computing platform. Notwithstanding these circumstances, our proposed network outperformed the comparison methods in terms of pair-searching speed. This result emphasizes the efficiency of our network.}